\documentclass[10pt,twocolumn,letterpaper]{article}

\usepackage[pagenumbers]{wacv} 

\newcommand{\red}[1]{{\color{red}#1}}

\definecolor{wacvblue}{rgb}{0.21,0.49,0.74}
\usepackage[pagebackref,breaklinks,colorlinks,allcolors=wacvblue]{hyperref}

\usepackage{amsmath}
\usepackage{booktabs}
\usepackage{multirow}
\usepackage{xcolor}
\usepackage{colortbl}              
\usepackage{etoolbox}
\usepackage{tabularray}

\usepackage[accsupp]{axessibility}  

\newcommand{\myparagraph}[1]{\vspace{1pt}\noindent{\bf #1}}
\newcommand{\blue}[1]{\textcolor{blue}{#1}}

\definecolor{mygray}{gray}{0.9} 

\def\wacvPaperID{2723} 
\def\confName{WACV}
\def\confYear{2026}

\title{Image-Guided Semantic Pseudo-LiDAR Point Generation \\ for 3D Object Detection}

\author{
Minseung Lee$^{1}$,\;
Seokha Moon$^{1}$,\;
Seung Joon Lee$^{2}$,\;
Reza Mahjourian$^{3}$,\;
and Jinkyu Kim$^{1\dagger}$\\[3pt]
$^{1}$CSE, Korea University \quad
$^{2}$LG Innotek \quad
$^{3}$Waymo Research\\[3pt]
\small{Correspondence: \texttt{jinkyukim@korea.ac.kr}}
}

\begin{document}

\maketitle

\renewcommand{\thefootnote}{\fnsymbol{footnote}} 
\footnotetext[2]{Corresponding author.}

\begin{abstract}
In autonomous driving scenarios, accurate perception is becoming an even more critical task for safe navigation. While LiDAR provides precise spatial data, its inherent sparsity makes it difficult to detect small or distant objects. Existing methods try to address this by generating additional points within a Region of Interest (RoI), but relying on LiDAR alone often leads to false positives and a failure to recover meaningful structures.
To address these limitations, we propose Image-Guided Semantic Pseudo-LiDAR Point Generation model, called ImagePG, a novel framework that leverages rich RGB image features to generate dense and semantically meaningful 3D points. Our framework includes an Image-Guided RoI Points Generation (IG-RPG) module, which creates pseudo-points guided by image features, and an Image-Aware Occupancy Prediction Network (I-OPN), which provides spatial priors to guide point placement. A multi-stage refinement (MR) module further enhances point quality and detection robustness. To the best of our knowledge, ImagePG is the first method to directly leverage image features for point generation. Extensive experiments on the KITTI and Waymo datasets demonstrate that ImagePG significantly improves the detection of small and distant objects like pedestrians and cyclists, reducing false positives by nearly 50\%. On the KITTI benchmark, our framework improves mAP by +1.38\%p (car), +7.91\%p (pedestrian), and +5.21\%p (cyclist) on the test set over the baseline, achieving state-of-the-art cyclist performance on the KITTI leaderboard. The code is available at: \href{https://github.com/MS-LIMA/ImagePG}{https://github.com/MS-LIMA/ImagePG}

\end{abstract}
\vspace{-1.5em}
\section{Introduction}

\begin{figure}[t!]
  \centering
  \includegraphics[width=1\linewidth]{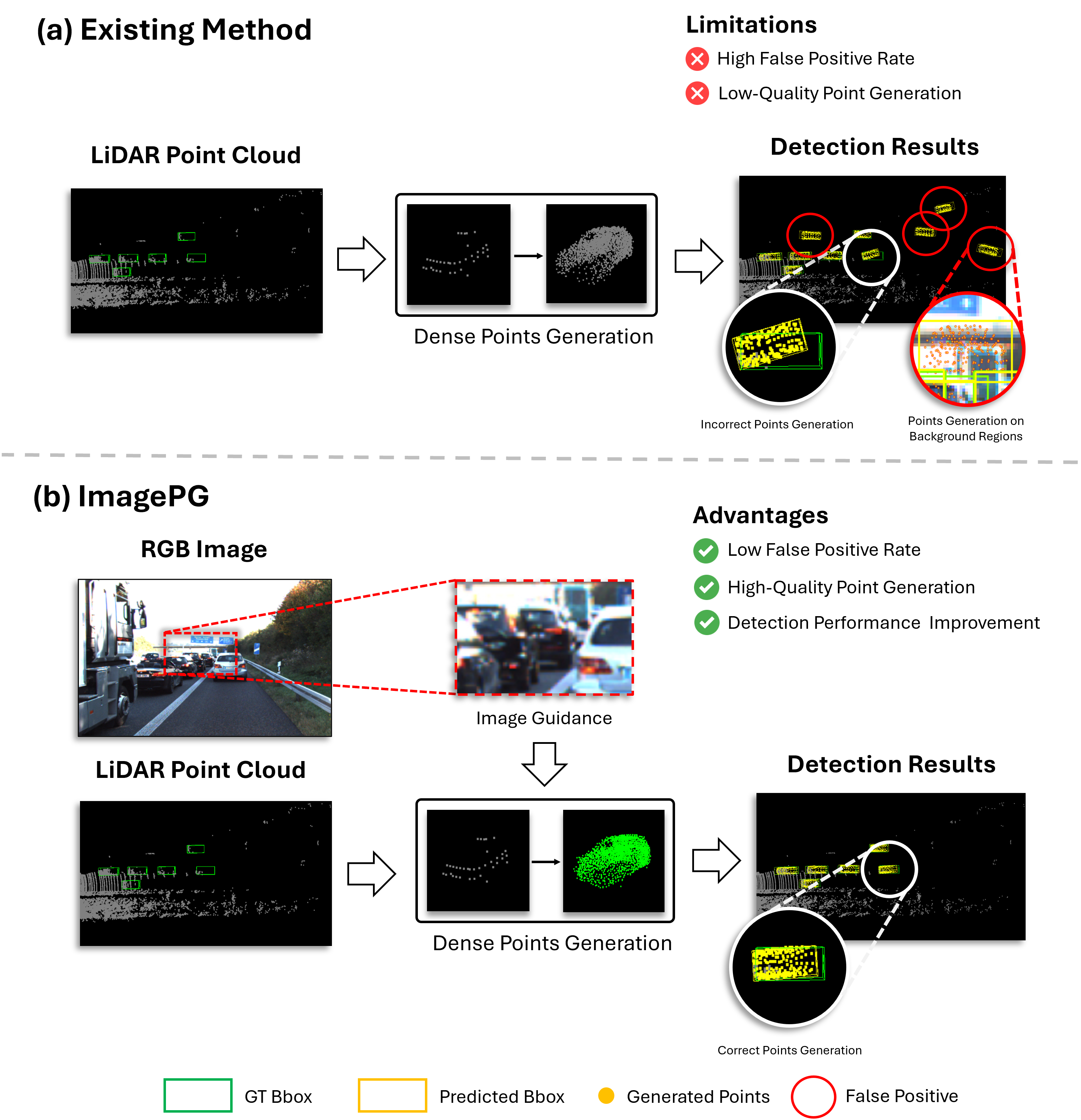}
  \captionof{figure}{(a) Conventional LiDAR-only methods for generating point clouds. (b) Our image-guided point cloud generation approach leverages visual semantic information (from RGB images) to enhance the density of point clouds, thereby improving overall 3D perception performance.
  }
  \label{fig:teaser}
  \vspace{-1.5em}
\end{figure}
In safety-critical applications like autonomous driving, robotics, and augmented reality, precise 3D object detection is a fundamental key task~\cite{kitti, augmented-reality, robotcleaning, voxelnet}. While LiDAR-based methods are the standard for 3D object detection due to their precise geometric information from point clouds~\cite{second, voxelnet, voxelrcnn, pvrcnn, pvrcnn++, pointrcnn, parta2}, they still struggle with small or distant objects, such as pedestrians and cyclists. These objects are often represented by only a handful of points, making robust localization and classification particularly challenging. The resulting sparse representations often yield incomplete geometry, which makes robust localization and classification particularly challenging~\cite{sienet, pcrgnn, pgrcnn, btcdet, htsspg, dsapg, vikienet, sfd}.

Recent methods have attempted to address the limitations of sparse LiDAR data by predicting object occupancy~\cite{spg, bshdet3d, btcdet}  or generating additional points to densify the input point clouds~\cite{sienet, pcrgnn, pgrcnn, htsspg, dsapg}, as shown in Figure~\ref{fig:teaser}(a). However, these approaches rely solely on LiDAR points to infer 3D occupancy or generate pseudo-points, which remains inherently challenging due to the limited density and coverage of LiDAR signals. This reliance often leads to inaccurate predictions and the misplacement of points in background areas. Our analysis shows that such methods frequently misclassify background regions as objects, resulting in a high false positive rate, while also failing to detect true objects because of insufficient data support. These limitations are especially critical in safety-sensitive domains such as autonomous driving, where detection failures or hallucinated objects can lead to incorrect planning and pose significant risks.

To address these problems, we propose Image-Guided Semantic Pseudo-LiDAR Point Generation (ImagePG), a novel framework that generates dense, semantically enriched 3D points by leveraging contextual cues from RGB images (Figure~\ref{fig:teaser}(b)). Our method comprises three modules—Image-Guided RoI Points Generation (IG-RPG), Image-Aware Occupancy Prediction Network (I-OPN), and multi-stage refinement (MR). IG-RPG leverages 2D visual features to supervise RoI-level pseudo-point generation, alleviating LiDAR sparsity and semantic ambiguity and yielding better localized, semantically aligned points. I-OPN predicts BEV-space occupancy to provide spatial priors that indirectly guide where points should be placed. MR iteratively refines both the generated points and the bounding boxes to enhance both point quality and detection accuracy. Together, our method explicitly generates semantic pseudo-LiDAR points under image supervision, yielding more informative point clouds.

To the best of our knowledge, ImagePG is the first method to directly leverage image features for point generation. Our experimental results demonstrate that the proposed method substantially improves 3D object detection performance, particularly for small and distant objects such as pedestrians and cyclists. On the KITTI~\cite{kitti} dataset, ImagePG yields substantial improvements of +12.01\%p (pedestrian) and +6.47\%p (cyclist) on the validation set, and +1.38\%p (car), +7.91\%p (pedestrian), +5.21\%p (cyclist) on the test set over the baseline~\cite{pgrcnn}, respectively, while nearly halving the number of false positives. Furthermore, ImagePG demonstrates generalization on the Waymo~\cite{waymo} dataset and across different backbones, suggesting its utility as a general-purpose enhancement module for 3D perception. Our key contributions are summarized as follows:

\vspace{0.3em}
\begin{itemize}
    \item We propose a novel image-guided 3D LiDAR point cloud generation method aimed at reliably enhancing 3D object detection performance, particularly for small and distant objects within a scene.
    \vspace{0.3em}
    \item To the best of our knowledge, we are the first to directly leverage image features for RoI-level point generation, markedly suppressing false positives caused by background completion, which is overlooked in prior works.
    \vspace{0.3em}
    \item We conduct extensive experiments, demonstrating its superiority over existing methods. Our results show that the method significantly improves the detection of sparse and distant objects, achieving state-of-the-art performance in cyclist detection on the KITTI leaderboard.
\end{itemize}
\vspace{-0.3em}
\section{Related Work}

\begin{figure*}[t!]
  \centering
  \includegraphics[width=0.95\linewidth]{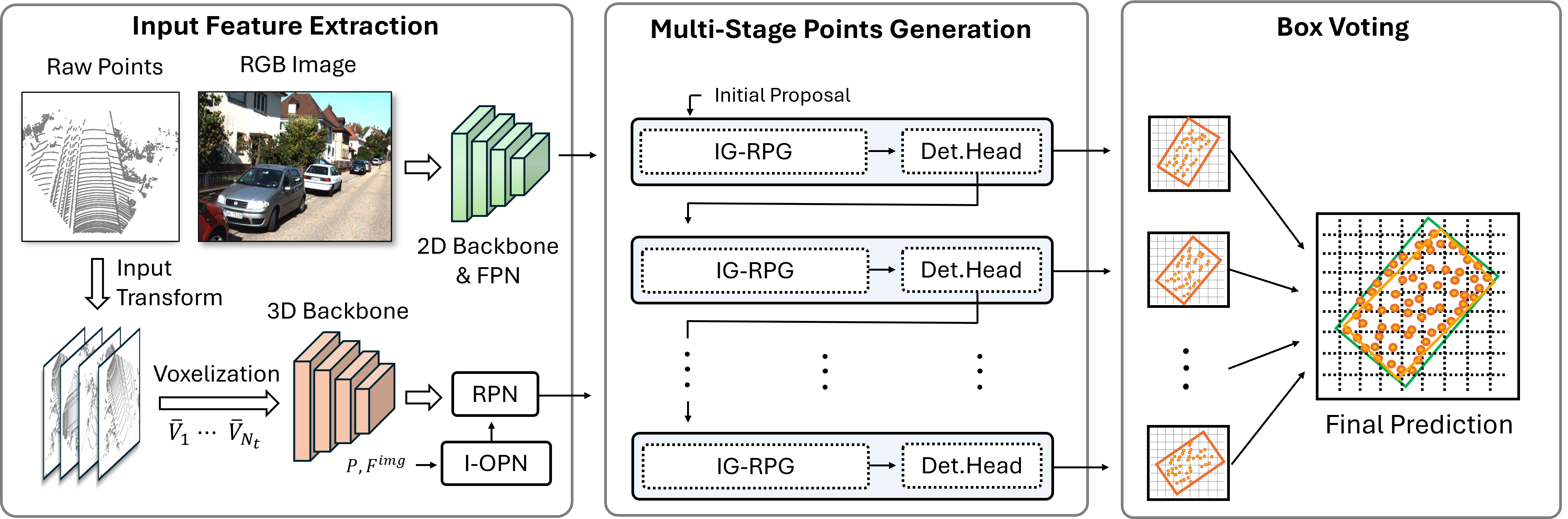}
  \vspace{-0.5em}
  \caption{An overview of our proposed ImagePG architecture. (i) The input point cloud undergoes multiple geometric transformations, and corresponding features are extracted alongside image features. (ii) Initial region proposals are generated by RPN in conjunction with I-OPN. (iii) These proposals are fed into the IG-RPG module to generate semantically enriched points, which are then used by the detection head to predict bounding boxes. The predicted boxes are passed to the next refinement stage. (iv) Final bounding boxes are obtained through a box voting mechanism that aggregates multi-stage predictions.}
  \label{fig:architecture}
\end{figure*}
\begin{figure*}[t!]
  \centering
  \includegraphics[width=0.95\linewidth]{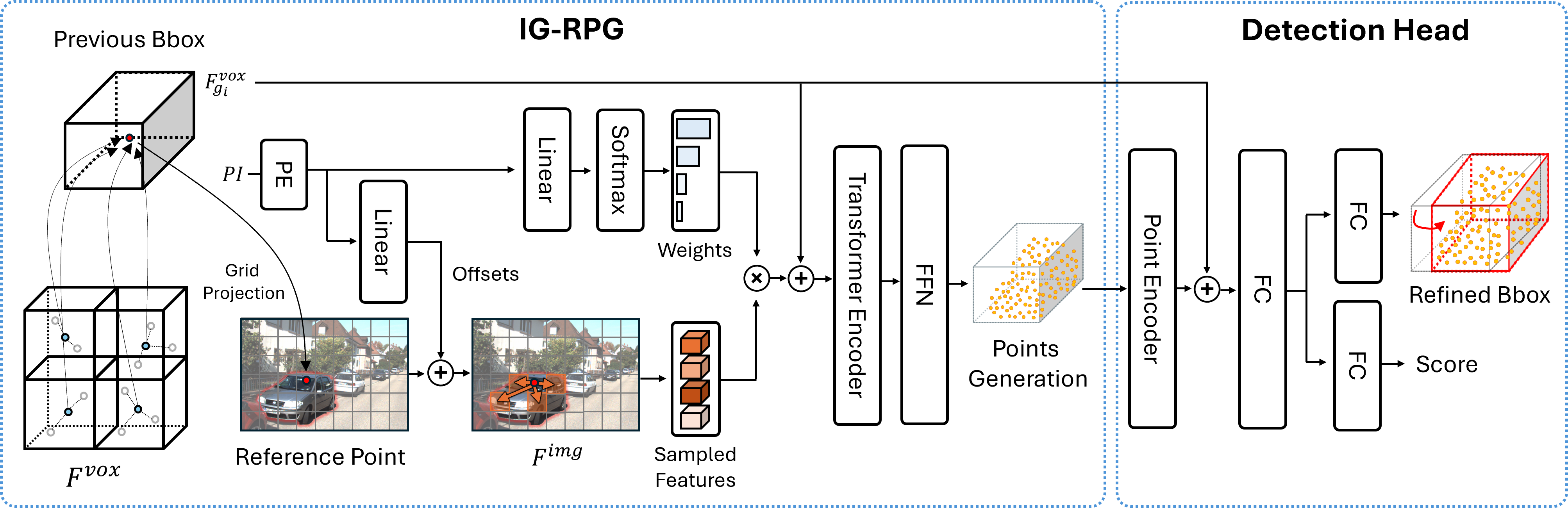}
  \vspace{-0.5em}
  \caption{Illustration of our proposed IG-RPG and the detection head.
(i) Each grid point is projected onto the image plane, where deformable attention is applied to sample corresponding image features. These sampled features are fused with the voxel features \(F_{g_i}^{vox}\), enabling the generation of semantically guided points. (ii) The generated points are subsequently encoded using a point encoder and passed to the detection head, which predicts 3D bounding boxes and classification scores.}
  \label{fig:ig_rpg}
  \vspace{-1em}
\end{figure*}

\myparagraph{LiDAR-based 3D Object Detection.}  
3D object detection involves localizing objects within a three-dimensional scene, typically represented as point cloud data. Point-based 3D object detectors extract features directly from raw points and detect 3D objects~\cite{pointnet, pointnet++, pointrcnn, 3dssd}. Grid-based methods address computational challenges by partitioning the 3D scene into a structured grid, such as voxels~\cite{voxelnet, voxelrcnn, second, pointpillars}. Some methods integrate both point-based and voxel-based techniques to leverage their advantages~\cite{pvrcnn, pvrcnn++, htsspg}. Recent approaches have improved detection performance by employing multiple refinement stages~\cite{ted, casa, virtualconv}. Nevertheless, LiDAR-based methods remain fundamentally limited by the inherent sparsity of LiDAR measurements.

\myparagraph{LiDAR-Camera 3D Object Detection.} 
In the domain of sensor fusion, early research efforts explored early fusion techniques by augmenting LiDAR points with semantic features from images~\cite{mvxnet, pointpainting, fusionpainting}. Other studies demonstrate significant improvements through feature-level fusion, where features from different modalities are concatenated rather than fused at the input stage~\cite{pointfusion, objectfusion, deepfusion, transfusion, 3dcvf, vff, logonet, epnet, focalsconv, upidet, catdet, graphalign}. Nevertheless, these methods neglect point sparsity during the fusion stage~\cite{vikienet}. In contrast, more recent approaches propose fusion strategies based on the bird’s-eye-view (BEV) space, utilizing view transform, such as LSS \cite{lss}, to project 2D image features into BEV space with dense semantic signals~\cite{bevfusion, isfusion, ealss, objectfusion, calico}. However, they fuse features from different modalities in a shared space, while neglecting the modality gap and misalignment between the two feature sets~\cite{vikienet}.

\myparagraph{Point Generation for 3D Object Detection.}  
Several studies~\cite{mvp, sfd, pseudolidar, pseudolidar++, virtualconv, ted, vikienet, sqd} augment LiDAR point clouds with pseudo points generated through heuristic methods or depth completion networks. However, these approaches often fail to generalize reliably due to errors and noises inherent in depth estimation, especially for small objects. Recent researches incorporate RoIs to generate denser point clouds, providing supplementary spatial supervision for 3D object detection~\cite{sienet, pgrcnn, htsspg, dsapg}. SIENet~\cite{sienet} uses a PCN-based completion network~\cite{pcn} to reconstruct missing structures, while PG-RCNN~\cite{pgrcnn} enhances RoI representations with generated points. However, these LiDAR-only methods can misclassify sparse background as foreground, yielding erroneous completions. In contrast, ImagePG integrates image semantics to generate meaningful points, thereby improving detection accuracy and reducing false positives.
\section{Method}

Our approach is designed around three complementary objectives. To mitigate points generation from the sparse point clouds, we propose the Image-Guided RoI Points Generation (IG-RPG) module, which fuses features from both voxel and image spaces to produce semantically enriched points. To improve the precision of both point generation and bounding box prediction, we incorporate a multi-stage refinement (MR) method that iteratively refines points generation and detection outputs, where TeSpConv~\cite{ted} is employed to capture transformation-equivariant features through multiple geometric transformations. Finally, to guide point generation with reliable spatial priors, we introduce the Image-Aware Occupancy Prediction Network (I-OPN), which predicts image-guided occupancy maps and thereby enhances overall detection performance.

\subsection{Multi-Stage Points Generation}
\myparagraph{Multi-Stage Refinement (MR).} As illustrated in Figure~\ref{fig:architecture}, the input point cloud \( P \) is first transformed via a set of operations \( T_{i=1}^{N_t} \), including rotation, reflection, and scaling, resulting in transformed point sets \( \bar{P}_{i=1}^{N_t} \), where \(N_t\) is the number of transformations. Each transformed point set is voxelized into \( \bar{V}_{i=1}^{N_t} \), which are then processed through a TeSpConv~\cite{ted, virtualconv} to extract voxel features \( \bar{F}_{i=1}^{N_t} \). These features are subsequently utilized across multiple refinement stages. This design leads to more accurate and semantically consistent pseudo points, enabling robustness to noise in any single transformation, as features are consolidated across diverse perspectives.

\begin{figure*}[t!]
  \centering
  \includegraphics[width=1\linewidth]{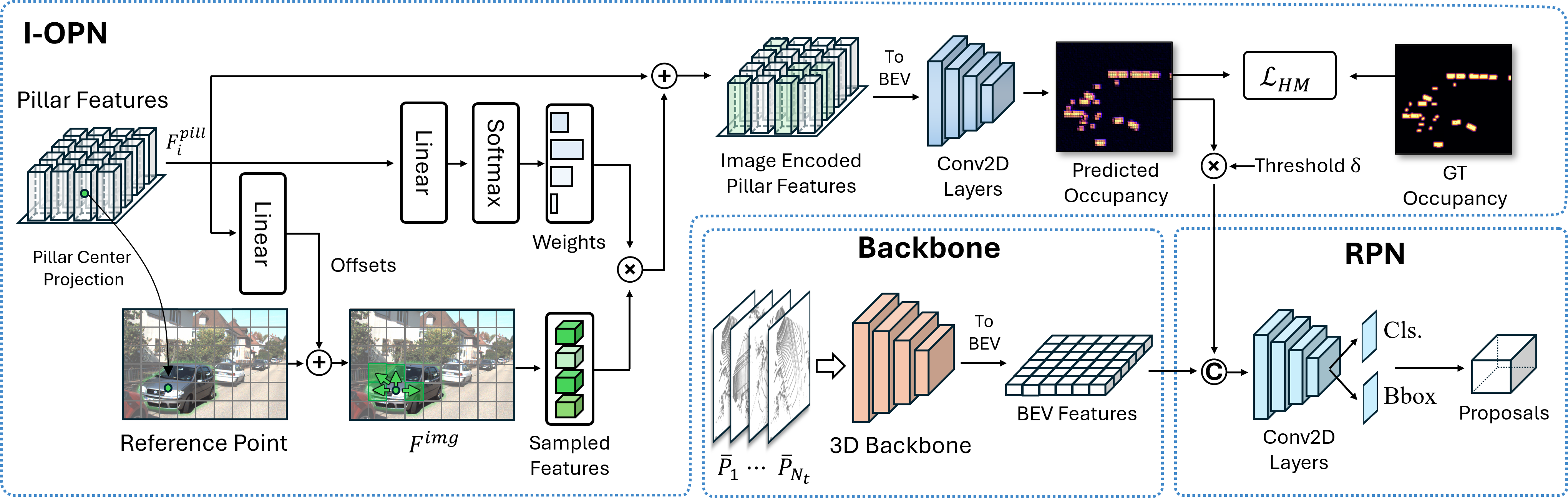}
  \vspace{-1.5em}
  \caption{Illustration of our proposed I-OPN. (i) Raw point clouds are encoded with a pillar-based backbone~\cite{pointpillars} to obtain pillar features \(F_i^{\text{pill}}\). Each pillar center is projected onto the image plane, deformable attention enriches \(F_i^{\text{pill}}\) with sampled image cues to yield \(F_i^{\text{spill}}\), and occupancy heatmap is predicted. (ii) From the multi-transformed LiDAR inputs, a shared 3D backbone extracts BEV features. (iii) The occupancy heatmap is concatenated with the BEV features, and the combined representation is fed into the RPN.}
  \label{fig:iopn}
  \vspace{-1.5em}
\end{figure*}

\myparagraph{Image-Guided RoI Points Generation (IG-RPG).} Previous methods~\cite{pgrcnn, dsapg, htsspg} rely solely on LiDAR points for point generation, often resulting in suboptimal accuracy and increased false positives. To address this, we propose Image-Guided RoI Points Generation (IG-RPG) module, a novel framework that produces points more semantically aligned with image features. Figure~\ref{fig:ig_rpg} illustrates IG-RPG module. Specifically, region proposals from the RPN are subdivided into \(G \times G \times G\) subvoxels, with their centers serving as grid points. Voxel features surrounding each grid point are aggregated via Voxel RoI Pooling~\cite{voxelrcnn}, yielding grid features \(F_{g_i}^{\text{vox}}\) from the input voxel features \(\bar{F}_{n_t}\) at the \(n_t\)-th refinement stage.

To capture the rich semantic information from RGB images, we utilize Deformable Attention~\cite{deformabledetr, logonet}, which extracts image features in relation to the corresponding grid points. Each grid point \( g_i \) is encoded using a Positional Encoding module \(\textit{PE}\) to generate positional embedding features \( F_{g_i}^{pe} \), which serve as queries. \(\textit{PE}\) encodes positional input \(\textit{PI}=[g_i - {r}^c; g_i - {r}^1; g_i - {r}^2; \cdots; g_i - {r}^8]\) with feed-forward network (FFN), where \({r}^c\) is the center and \({r}^{1,2,\cdots,8}\) are the eight corners of the 3D bounding box. The grid point \( g_i \) is then projected onto the image plane, denoted as \( g_{pi} \). Then, grid image feature $F^\text{img}_{g_i}$ can be extracted by \(\texttt{DeformAttn}(F^{pe}_{g_i}, g_{pi}, F^\text{img})\), where $F^{img}$ is from the image backbone. 

Fused grid feature can be calculated as: $\tilde{F}_{g_i}^{\text{fus}} = \mathcal{T}(F_{g_i}^\text{vox}+F^\text{img}_{g_i}, \delta_{g_i})$, where \(\mathcal{T}\) is the Transformer encoder \cite{transformer}. Positional encoding is generated with feed-forward network (FFN) and can be formulated as  
\(\delta_{g_i} = \texttt{FFN}(\textit{PI})\). Finally, MLPs generate offsets and semantic features of points from the center of the grid \(g_i\): i.e., $[{o}_i; F^{\text{sem}}_{{p}_i}] = \texttt{MLP}_{gen}(\tilde{F}_{{g}_i}^{\text{fus}})$. The offsets of the points from the grid centers are generated, therefore, actual point’s coordinates \({p}_i = (x_i, y_i, z_i)\) are calculated as \({g}_i + {o}_i\). In addition, foreground scores \(s_i\) for generated points are calculated by MLPs and a sigmoid function \(\sigma\), which can be written as ${s}_i = \sigma(\texttt{MLP}_s(F^{{\text{sem}}}_{{p}_i}))$.

To optimize, we minimize \(\mathcal{L}_{\text{RPG}}\) = \(\mathcal{L}_{\text{score}}\) + \(\mathcal{L}_{\text{offset}}\). \(\mathcal{L}_{\text{offset}}\) ensures that the generated points are geometrically closer to the complete points of objects. We use Chamfer distance described as follows:
\begin{equation}
\begin{aligned}
    \mathcal{L}_{\text{offset}} &= \frac{1}{N_{p}} \sum_{r} \Bigg( 
        \frac{1}{|\mathbf{S}_1|} \sum_{\mathbf{x} \in \mathbf{S}_1} 
        \min_{\mathbf{y} \in \mathbf{S}_2} \|\mathbf{x} - \mathbf{y}\|_2^2 \\
    &\quad + \frac{1}{|\mathbf{S}_2|} \sum_{\mathbf{y} \in \mathbf{S}_2} 
        \min_{\mathbf{x} \in \mathbf{S}_1} \|\mathbf{y} - \mathbf{x}\|_2^2 
    \Bigg),
\end{aligned}
\end{equation}

\noindent where \(N_{p}\) is the number of positive proposals, and \(\mathbf{S}_1\) and \(\mathbf{S}_2\) represent the count of generated points and the dense ground truth (GT) object points of the \(r\)-th foreground proposal, \(r = 1, 2, \cdots, N_{p}\). Further, \(\mathcal{L}_{\text{score}}\) ensures that the generated points reside within the GT bounding boxes. The FPS algorithm \cite{pointnet++} is used to sample the generated points and reduce computational costs. \(\mathcal{L}_{\text{score}}\) can be described as follows:
\begin{equation}
    \mathcal{L}_{\text{score}} = -\frac{1}{N_s} \sum_j \left( (1 - s_j)^\gamma \log s_j \right),
\end{equation}
where \(N_s\) is the number of sampled points.

\noindent\textbf{Detection Head.}
Following~\cite{pgrcnn, parta2, pointrcnn}, generated points are first transformed into the canonical coordinate system based on region proposals. We incorporate depth, computed as \( d_i = \sqrt{x_i^2 + y_i^2 + z_i^2} \), to provide global context, and append the point-wise score \( s_i \) to reflect semantic importance. The final canonical features are extracted using an MLP as \( F^{\text{can}}_{p_i} = \texttt{MLP}_{\text{can}}([x_i^c, y_i^c, z_i^c, d_i, s_i]) \), where \( (x_i^c, y_i^c, z_i^c) \) are canonical coordinates. Lastly, PointNet++ \cite{pointnet++} is used to extract comprehensive features for the RoI, which are described as \(F^{\text{roi}} = \texttt{PointEncoder}\left(\left[p_i;F^{\text{can}}_{{p}_i};F^{\text{sem}}_{p_i}\right]\right) + F_{g_i}^\text{vox}.\) Then, the detection head produces the final bounding box predictions and confidence scores. Predicted boxes will serve as region proposals for the next stage of refinement. We adopt weighted box voting \cite{ted, casa, virtualconv} to produce final refined bounding boxes and confidence scores.

\vspace{-0.4em}
\subsection{Image-Aware Occupancy Prediction Network}
Occupancy prediction networks (OPN) improve 3D perception by enhancing proposal generation and downstream detection~\cite{3dhanet, htsspg, bshdet3d}, however, LiDAR-only OPNs often yield suboptimal results, similar to LiDAR-only point generation. Nevertheless, discarding occupancy prediction is not an option, as ignoring occupancy risks generating points in empty regions and can increase false positives. To this end, we propose Image-Aware Occupancy Prediction Network (I-OPN), which jointly optimizes occupancy prediction. By explicitly predicting occupancy, I-OPN guides pseudo-point generation toward semantically meaningful and geometrically plausible regions, thereby improving detection performance.

As shown in Figure~\ref{fig:iopn}, the input point cloud \(P\) is encoded into pillar features \(F_i^{\text{pill}}\) using a pillar-based network~\cite{pointpillars}, and each pillar center is projected onto the image plane. Image-guided features \(F_i^{\text{spill}} = \texttt{DeformAttn}(F_i^{\text{pill}}, pill_{pi}, F^{\text{img}})\) are compressed along the vertical axis to obtain BEV features, where \(pill_{pi}\) is the projected pillar center. These features are passed through convolutional layers~\cite{second} to predict an occupancy heatmap. The loss function can be formulated as : 
\vspace{-0.2em}
\begin{equation}
    \mathcal{L}_{\text{HM}} = -(1 - q_t)^\gamma \log(q_t), 
\end{equation}

\noindent where $q_t$ represents the probability of occupancy for each grid on the BEV feature map. The occupancy prediction is thresholded using \(\delta = 0.5\), and the resulting binary map is concatenated with multi-transformation BEV features before being fed into the RPN. 

\vspace{-0.4em}
\subsection{Loss Function}
Concretely, we minimize the following loss function $\mathcal{L}_{\text{total}}$:
\vspace{-0.2em}
\begin{equation}
    \mathcal{L}_{\text{total}} = \mathcal{L}_{\text{RPN}} + \mathcal{L}_{\text{RCNN}} + \mathcal{L}_{\text{RPG}} + \mathcal{L}_{\text{HM}}
\end{equation}
where \(\mathcal{L}_{\text{RPN}}\) is the RPN loss proposed by \cite{second}, \(\mathcal{L}_{\text{RCNN}}\) is the proposal refinement loss from \cite{voxelrcnn}, \(\mathcal{L}_{\text{RPG}}\) is the point generation loss from IG-RPG, and \(\mathcal{L}_{\text{HM}}\) is from the I-OPN. \(\mathcal{L}_{\text{RPN}}\) and \(\mathcal{L}_{\text{RCNN}}\) are fundamentally similar, predicting RoIs' classes and bounding boxes. For bounding box regression, only positive predictions that exceed the intersection-over-union (IoU) threshold participate in loss calculation using the smooth-L1 loss. For class classification, focal loss \cite{focalloss} is used for the RPN, and binary cross-entropy is used for the detection head.

\section{Experiments}

\subsection{Experimental Setup}

\begin{figure}[t!]
  \centering
  \includegraphics[width=1\linewidth]{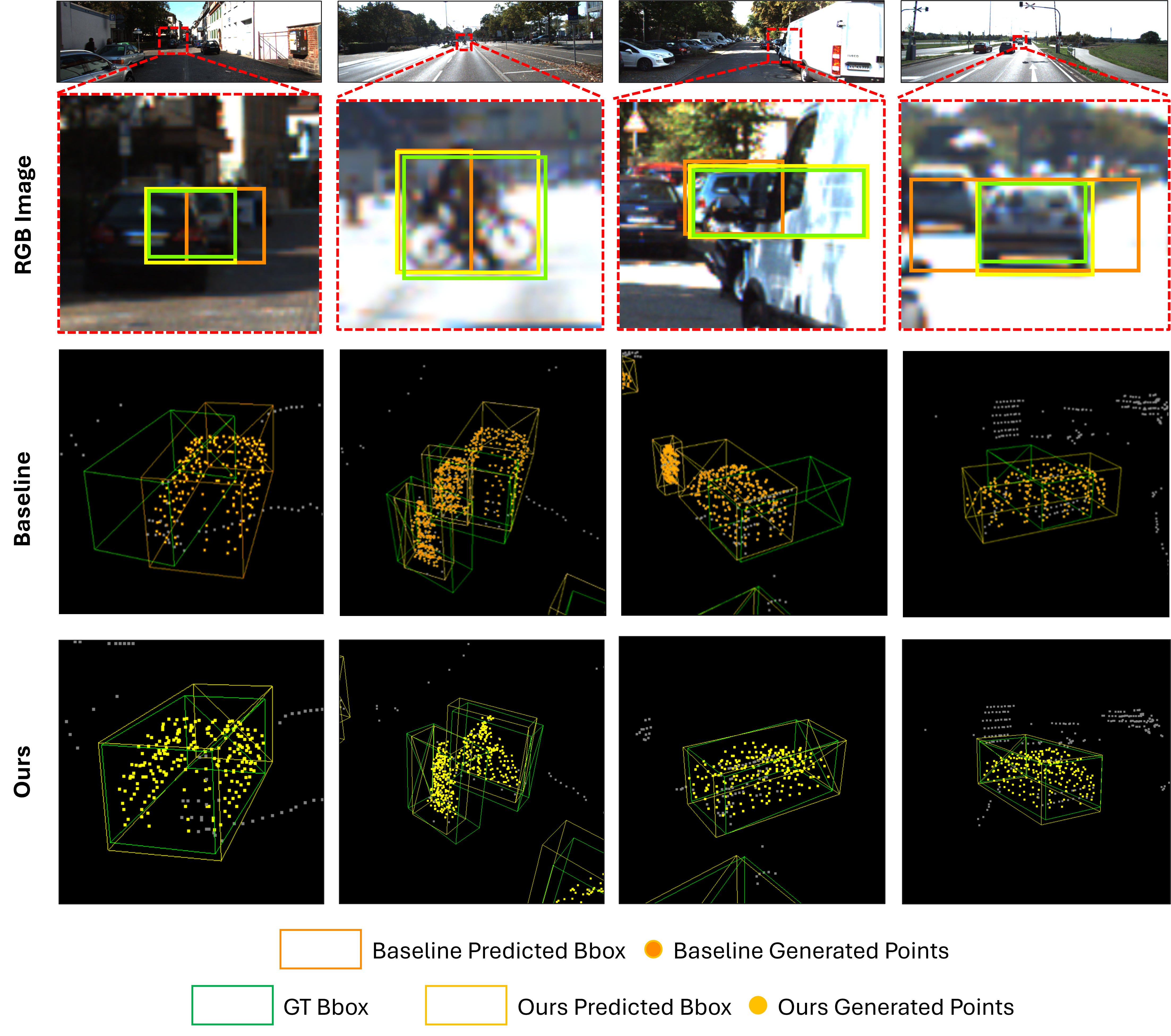}
  \vspace{-1.5em}
  \caption{Qualitative results of semantic point generation and detection for baseline~\cite{pgrcnn} and ours. The baseline~\cite{pgrcnn} often generates incorrect points, whereas our approach suppresses such errors, leading to improved detection performance.}
  \label{fig:qual_point_generation}
  \vspace{-1.5em}
\end{figure}

\renewcommand{\arraystretch}{0.8}

\begin{table*}[t!]
\vspace{1.0em}
\centering
\scriptsize
\setlength{\tabcolsep}{4.5pt}
{
\begin{tabular}{l||c|c|c c c c|c c c c|c c c c}
\toprule
    \multirow{2}{*}{\textbf{Method}} & \multirow{2}{*}{\textbf{Modality}} & \multirow{2}{*}{\textbf{mAP}}& \multicolumn{4}{c|}{\textbf{Car 3D} \scriptsize{(IoU=0.7)}} & \multicolumn{4}{c|}{\textbf{Pedestrian 3D} \scriptsize{(IoU=0.5)}} & \multicolumn{4}{c}{\textbf{Cyclist 3D} \scriptsize{(IoU=0.5)}} \\
    \cline{4-15} 
    \rule{0in}{1.25em} & & & \textbf{Easy} & \textbf{Mod.} & \textbf{Hard} & \textbf{mAP} & \textbf{Easy} & \textbf{Mod.} & \textbf{Hard} & \textbf{mAP} & \textbf{Easy} & \textbf{Mod.} & \textbf{Hard} & \textbf{mAP} \\
\midrule
        SECOND \cite{second} & L & 68.05 & 90.68 & 79.02 & 75.39 & 81.70 & 59.23 & 52.35 & 45.98 & 52.52 & 83.03 & 65.35 & 61.39 & 69.92 \\
        PointPillars \cite{pointpillars} & L & 67.12 & 88.26 & 78.90 & 76.06 & 81.07 & 57.10 & 50.96 & 46.38 & 51.48 & 83.77 & 62.99 & 59.65 & 68.80 \\
        PointRCNN \cite{pointrcnn} & L & 73.63 & 89.13 & 78.72 & 78.24 & 82.03 & 65.81 & 59.57 & 52.75 & 59.38 & 93.51 & 74.19 & 70.73 & 79.48 \\
        PV-RCNN \cite{pvrcnn} & L & 73.98 & 91.84 & 82.40 & 78.84 & 84.36 & 66.21 & 58.92 & 53.77 & 59.63 & 90.45 & 73.30 & 69.38 & 77.71 \\
        Voxel R-CNN \cite{voxelrcnn} & L & - & 91.72 & 83.19 & 78.60 & 84.50 & - & - & - & - & - & - & - & - \\
        PC-RGNN \cite{pcrgnn} & L & - & 90.94 & 81.43 & 80.45 & 84.27 & - & - & - & - & - & - & - & - \\
        SIENet \cite{sienet} & L & - & 92.49 & 85.43 & 83.05 & 86.99 & - & - & - & - & - & - & - & - \\
        PDV \cite{pdv} & L & 75.67 & 92.56 & 85.29 & 83.05 & 86.97 & 66.90 & 60.80 & 55.85 & 61.18 & 92.72 & 74.23 & 69.60 & 78.85 \\
        PG-RCNN \cite{pgrcnn} & L & 76.01 & 92.73 & 85.26 & 82.83 & 86.94 & 68.44 & 60.63 & 55.36 & 61.48 & 93.84 & 74.85 & 70.15 & 79.61 \\
        CasA-V \cite{casa} & L & 77.80 & 93.21 & 86.37 & 83.94 & 87.84 & 73.95 & 66.62 & 59.97 & 66.85 & 92.78 & 73.94 & 69.37 & 78.70 \\   
        TED-S \cite{ted} & L & 78.95 & 93.05 & 87.91 & 85.81 & 88.92 & 72.38 & 67.81 & 63.54 & 67.91 & 93.09 & 75.77 & 71.20 & 80.02 \\
        HT-SSPG \cite{htsspg} & L & 79.38 & 93.16 & 86.06 & 83.51 & 87.57 & \underline{75.95} & 69.09 & 63.44 & 69.49 & \underline{95.86} & 76.04 & 71.32 & 81.07 \\
        DSaPG \cite{dsapg} & L & 77.76 & 93.08 & 85.30 & 83.24 & 87.21 & 72.07 & 64.34 & 59.75 & 65.39 & 94.19 & 75.20 & \underline{72.65} & 80.68 \\
\midrule
        EPNet \cite{epnet} & L+C & 70.81 & 88.76 & 78.65 & 78.32 & 81.91 & 66.74 & 59.29 & 54.82 & 60.28 & 83.88 & 65.60 & 62.70 & 70.23 \\
        FocalsConv \cite{focalsconv} & L+C & - & 92.26 & 85.32 & 82.95 & 87.01 & - & - & - & - & - & - & - & - \\
        CAT-Det \cite{catdet} & L+C & 75.42 & 90.12 & 81.46 & 79.15 & 83.58 & 74.08 & 66.35 & 58.92 & 66.45 & 87.64 & 72.82 & 68.20 & 76.22 \\
        VFF \cite{vff} & L+C & 77.14 & 92.31 & 85.82 & 82.92 & 87.72 & 73.26 & 65.11 & 60.03 & 66.25 & 89.40 & 73.12 & 69.86 & 77.46 \\
        SFD \cite{vff} & L+C & 78.24 & \textbf{95.52} & 88.27 & 85.57 & 89.79 & 72.94 & 66.69 & 61.59 & 67.07 & 93.39 & 72.95 & 67.26 & 77.87 \\
        LoGoNet \cite{logonet} & L+C & 77.11 & 92.04 & 85.04 & 84.31 & 87.02 & 70.20 & 63.72 & 59.46 & 64.46 & 91.74 & 75.35 & 72.42 & 79.84 \\
        VirConv-T \cite{virtualconv} & L+C & 78.40 & \underline{94.98} & \textbf{89.96} & \textbf{88.13} & \textbf{90.65} & 73.32 & 66.93 & 60.38 & 66.88 & 90.04 & 73.90 & 69.06 & 77.67 \\
        GraphAlign \cite{graphalign} & L+C & - & 92.44 & 87.01 & 84.68 & 88.04 & - & - & - & - & - & - & - & - \\  
        UPIDet \cite{upidet} & L+C & 79.43 & 92.42 & 86.03 & 83.49 & 87.31 & 75.45 & \underline{69.79} & \underline{63.69} & \underline{69.64} & 95.17 & \underline{76.75} & 72.12 & \underline{81.35} \\
        TED-M \cite{ted} & L+C & \underline{80.15} & 95.25 & \underline{88.94} & \underline{86.73} & \underline{90.31} & 74.73 & 69.07 & 63.63 & 69.14 & 95.20 & 76.17 & 71.59 & 80.99 \\
        ViKIENet \cite{vikienet} & L+C & 78.13 & 92.99 & 87.63 & 85.48 & 88.70 & 75.35 & 67.15 & 61.80 & 68.10 & 90.88 & 72.87 & 69.06 & 77.60 \\
\midrule
       \rowcolor{gray!10} Baseline$^\dagger$ \cite{pgrcnn} & L & 75.31 & 92.49 & 84.87 & 82.40 & 86.59 & 65.86 & 58.85 & 53.47 & 59.39 & 91.29 & 72.13 & 67.44 & 76.95 \\
        \rowcolor{gray!10}+ ImagePG (Ours) & L+C & \textbf{80.66} & 92.41 & 85.67 & 83.45 & 87.18 & \textbf{77.64} & \textbf{71.01} & \textbf{65.56} & \textbf{71.40} & \textbf{96.53} & \textbf{78.94} & \textbf{74.79} & \textbf{83.42} \\
        \rowcolor{gray!10} {\tiny \textit{}} & & {\tiny (\red{5.35\%$\uparrow$})} & {\tiny 
 (\blue{0.08\%$\downarrow$})} & {\tiny (\red{0.80\%$\uparrow$})} & {\tiny (\red{1.05\%$\uparrow$})} & {\tiny (\red{0.59\%$\uparrow$})} & {\tiny (\red{11.78\%$\uparrow$})} & {\tiny (\red{12.16\%$\uparrow$})} & {\tiny (\red{12.09\%$\uparrow$})} & {\tiny (\red{12.01\%$\uparrow$})} & {\tiny (\red{5.24\%$\uparrow$})} & {\tiny (\red{6.81\%$\uparrow$})} & {\tiny (\red{7.35\%$\uparrow$})} & {\tiny (\red{6.47\%$\uparrow$})} \\
\bottomrule
\end{tabular}
}
\vspace{-0.5em}
\caption{3D detection performance comparison on the KITTI~\cite{kitti} \textit{val} set calculated by 40 recall positions. Note that the highest score is marked with \textbf{bold} and the second with \underline{underline}. L and C denote the LiDAR point cloud and the image, respectively. $^\dagger$: reproduced.}
\label{tab:kitti_val}
\vspace{-0.5em}
\end{table*}
\renewcommand{\arraystretch}{0.8}

\begin{table*}[t!]
\centering
\scriptsize
\setlength{\tabcolsep}{4.9pt}
{
\begin{tabular}{l||c|c|c c c c|c c c c|c c c c}
\toprule
    \multirow{2}{*}{\textbf{Method}} & \multirow{2}{*}{\textbf{Modality}} & \multirow{2}{*}{\textbf{mAP}}& \multicolumn{4}{c|}{\textbf{Car 3D} \scriptsize{(IoU=0.7)}} & \multicolumn{4}{c|}{\textbf{Pedestrian 3D} \scriptsize{(IoU=0.5)}} & \multicolumn{4}{c}{\textbf{Cyclist 3D} \scriptsize{(IoU=0.5)}} \\
    \cline{4-15} 
    \rule{0in}{1.25em} & & & \textbf{Easy} & \textbf{Mod.} & \textbf{Hard} & \textbf{mAP} & \textbf{Easy} & \textbf{Mod.} & \textbf{Hard} & \textbf{mAP} & \textbf{Easy} & \textbf{Mod.} & \textbf{Hard} & \textbf{mAP} \\
\midrule
        PointRCNN \cite{pointrcnn} & L & 60.33 & 86.96 & 75.64 & 70.70 & 77.77 & 47.98 & 39.37 & 36.01 & 41.12 & 74.96 & 58.82 & 52.53 & 62.10 \\
        PV-RCNN \cite{pvrcnn} & L & 64.91 & 90.25 & 81.43 & 76.82 & 82.83 & 52.17 & 43.29 & 40.29 & 45.25 & 78.60 & 63.71 & 57.65 & 66.65 \\
        PDV \cite{pdv} & L & 65.31 & 90.43 & 81.86 & 77.36 & 83.21 & 47.80 & 40.56 & 38.46 & 42.27 & 83.04 & 67.81 & 60.46 & 70.44 \\
        PG-RCNN \cite{pgrcnn} & L & 65.38 & 89.38 & 82.13 & 77.33 & 82.95 & 47.99 & 41.04 & 38.71 & 42.58 & 82.77 & 67.82 & 61.25 & 70.61 \\
        CasA-V \cite{casa} & L & 69.77 & 91.58 & 83.06 & 80.08 & 84.91 & 54.04 & 47.09 & 44.56 & 48.56 & \underline{87.91} & 73.47 & 66.17 & 75.85 \\   
\midrule
        CAT-Det \cite{catdet} & L+C & 67.05 & 89.87 & 81.32 & 76.68 & 82.62 & 54.26 & 45.44 & 41.94 & 47.21 & 83.68 & 68.81 & 61.45 & 71.31 \\
        LoGoNet \cite{logonet} & L+C & 69.35 & \textbf{91.80} & \underline{85.06} & \textbf{80.74} & \textbf{85.87 }& 53.07 & 47.43 & 45.22 & 48.57 & 84.47 & 71.70 & 64.67 & 73.61 \\
        GraphAlign \cite{graphalign} & L+C & 63.07 & 90.90 & 82.23 & 79.67 & 84.27 & 41.38 & 36.89 & 34.95 & 37.74 & 78.42 & 64.43 & 58.71 & 67.19 \\  
        UPIDet \cite{upidet} & L+C & 70.13 & 89.13 & 82.97 & 80.05 & 84.05 & 55.59 & 48.77 & \underline{46.12} & 50.16 & 86.74 & \underline{74.32} & \textbf{67.45} & \underline{76.17} \\
        TED-M \cite{ted} & L+C & \textbf{70.99} & \underline{91.61} & \textbf{85.28} & \underline{80.68} & \underline{85.86} & \underline{55.85} & \textbf{49.21} & \textbf{46.52} & \textbf{50.53} & \textbf{88.82} & 74.12 & \underline{66.84} & \textbf{76.59} \\
\midrule
        \rowcolor{gray!10} Baseline \cite{pgrcnn} & L & 65.38 & 89.38 & 82.13 & 77.33 & 82.95 & 47.99 & 41.04 & 38.71 & 42.58 & 82.77 & 67.82 & 61.25 & 70.61 \\
        \rowcolor{gray!10}+ ImagePG (Ours) & L+C & \underline{70.21} & 91.00 & 82.46 & 79.54 & 84.33 & \textbf{56.66} & \underline{48.81} & 45.99 & \underline{50.49} & 86.53 & \textbf{74.68} & 66.24 & 75.82 \\
        \rowcolor{gray!10} {\tiny \textit{}} & & {\tiny (\red{4.83\%$\uparrow$})} & {\tiny (\red{1.62\%$\uparrow$})} & {\tiny (\red{0.33\%$\uparrow$})} & {\tiny (\red{2.21\%$\uparrow$})} & {\tiny (\red{1.38\%$\uparrow$})} & {\tiny (\red{8.67\%$\uparrow$})} & {\tiny (\red{7.77\%$\uparrow$})} & {\tiny (\red{7.28\%$\uparrow$})} & {\tiny (\red{7.91\%$\uparrow$})} & {\tiny (\red{3.76\%$\uparrow$})} & {\tiny (\red{6.86\%$\uparrow$})} & {\tiny (\red{4.99\%$\uparrow$})} & {\tiny (\red{5.21\%$\uparrow$})} \\
\bottomrule
\end{tabular}
}
\vspace{-0.5em}
\caption{3D detection performance comparison on the KITTI~\cite{kitti} \textit{test} set calculated by 40 recall positions. Highest result is marked with \textbf{bold} and second with \underline{underline}.}
\label{tab:kitti_test}
\vspace{-1.5em}
\end{table*}

\myparagraph{Datasets.} 
We conduct experiments using the KITTI dataset~\cite{kitti}, which includes 7,481 training samples and 7,518 test samples across three object classes: car, pedestrian, and cyclist. Following standard evaluation protocol~\cite{kitti}, the training set is further divided into 3,712 training and 3,769 validation samples. In addition, we also evaluate our model with the Waymo Open Dataset (WOD)~\cite{waymo}, a large-scale autonomous driving dataset comprising 798 training sequences and 202 validation sequences for the three classes (i.e., vehicle, pedestrian, and cyclist). Note that the dataset was collected using a sensor suite that includes five surround-view cameras. Additional dataset details are provided in the supplemental material.

\myparagraph{Implementation Details.} For the KITTI~\cite{kitti} dataset, the detection range is set to [0m, 70.4m] along the X-axis, [-40m, 40m] along the Y-axis, and [-3m, 1m] along the Z-axis, as it provides annotations only for objects visible in the camera images. The voxel size is set to (0.05m, 0.05m, 0.05m). Random flipping, global scaling, and global rotations are used for data augmentation. Following the conventions, we also adopt ground truth (GT) sampling~\cite{second, mixaug}.

For the Waymo~\cite{waymo} dataset, the detection range is set to [-75.2m, 75.2m] along both the X-axis and Y-axis, and [-2m, 4m] along the Z-axis. The voxel size is set to (0.1m, 0.1m, 0.15m). We employed the same global rotation and scaling augmentation strategies as described for the KITTI~\cite{kitti} dataset, including flipping along the X and Y-axes. Following baseline's settings~\cite{pgrcnn}, we used 20\% of the train dataset, and images are downsampled by 2$\times$. More detailed implementation details will be provided in the supplemental material. 

\myparagraph{Evaluation Details.} 
For the KITTI~\cite{kitti} dataset, we evaluated our method using mean average precision (mAP) for 40 recall positions (R40), with IoU thresholds set to 0.7, 0.5, and 0.5 for cars, pedestrians, and cyclists, respectively. For the Waymo~\cite{waymo} dataset, mAPH (mean average precision by heading) is also used and our evaluation was conducted for all difficulty levels: i.e., Level 1 (L1) and 2 (L2), where L1 refers to ground truth boxes with more than 5 points, and L2 refers to boxes with fewer or equal to 5 points.

\renewcommand{\arraystretch}{0.7}
\begin{table*}[t!]
\vspace{1.0em}
\centering
\scriptsize
\setlength{\tabcolsep}{7.75pt}
{
\begin{tabular}{l||cc|cc|cc|cc|cc|cc}
\toprule
    \multirow{3}{*}{\textbf{Method}} 
    & \multicolumn{4}{c|}{\textbf{Vehicle 3D} \scriptsize{(IoU=0.7)}} 
    & \multicolumn{4}{c|}{\textbf{Pedestrian 3D} \scriptsize{(IoU=0.5)}} 
    & \multicolumn{4}{c}{\textbf{Cyclist 3D} \scriptsize{(IoU=0.5)}} \\
\cline{2-13}
    \rule{0pt}{1.25em} & \multicolumn{2}{c|}{\textbf{L1}} & \multicolumn{2}{c|}{\textbf{L2}}
    & \multicolumn{2}{c|}{\textbf{L1}} & \multicolumn{2}{c|}{\textbf{L2}}
    & \multicolumn{2}{c|}{\textbf{L1}} & \multicolumn{2}{c}{\textbf{L2}} \\
\cline{2-13}
    \rule{0pt}{1.25em} & \textbf{AP} & \textbf{APH} & \textbf{AP} & \textbf{APH} 
    & \textbf{AP} & \textbf{APH} & \textbf{AP} & \textbf{APH} 
    & \textbf{AP} & \textbf{APH} & \textbf{AP} & \textbf{APH} \\
\midrule
    Baseline\(^\dagger\)~\cite{pgrcnn}
    & 75.00 & 74.52 & 66.49 & 66.06
    & 75.14 & 68.79 & 66.19 & 60.44
    & 69.82 & 68.74 & 67.23 & 66.19 \\
\rowcolor{gray!10} \rule{0in}{1em} + ImagePG (Ours)
    & 77.31 & 76.85 & 68.84 & 68.41
    & 78.11 & 72.53 & 69.28 & 64.14
    & 71.31 & 70.29 & 68.71 & 67.73 \\
    
    \rowcolor{gray!10}
    \tiny 
    & \tiny (\textcolor{red}{2.31\%↑}) & \tiny (\textcolor{red}{2.33\%↑}) & \tiny (\textcolor{red}{2.35\%↑}) & \tiny (\textcolor{red}{2.35\%↑})
    & \tiny (\textcolor{red}{2.97\%↑}) & \tiny (\textcolor{red}{3.74\%↑}) & \tiny (\textcolor{red}{3.09\%↑}) & \tiny (\textcolor{red}{3.70\%↑})
    & \tiny (\textcolor{red}{1.49\%↑}) & \tiny (\textcolor{red}{1.55\%↑}) & \tiny (\textcolor{red}{1.48\%↑}) & \tiny (\textcolor{red}{1.54\%↑}) \\
\bottomrule
\end{tabular}
}
\vspace{-0.75em}
\caption{3D detection performance comparison on the Waymo \cite{waymo} \textit{val} set for vehicle, pedestrian and cyclist classes. $^\dagger$: reproduced.}
\label{tab:waymo_val}
\vspace{-0.5em}
\end{table*}

\renewcommand{\arraystretch}{0.7}
\setlength{\tabcolsep}{12pt}
\begin{table*}[t!]
\centering
\scriptsize
{
\begin{tabular}{l||c c c|c c c|c c c}
\toprule
    \multirow{2}{*}{\textbf{Method}} & \multicolumn{3}{c|}{\textbf{Car 3D} \scriptsize{(IoU=0.7)}} & \multicolumn{3}{c|}{\textbf{Pedestrian 3D} \scriptsize{(IoU=0.5)}} & \multicolumn{3}{c}{\textbf{Cyclist 3D} \scriptsize{(IoU=0.5)}} \\
    \cline{2-10}
\rule{0pt}{1.25em}
& \textbf{[0, 20)} & \textbf{[20, 40)} & \textbf{[40, inf)} 
& \textbf{[0, 20)} & \textbf{[20, 40)} & \textbf{[40, inf)} 
& \textbf{[0, 20)} & \textbf{[20, 40)} & \textbf{[40, inf)} \\
\midrule
Baseline\(^\dagger\)~\cite{pgrcnn} & 95.12 & 84.40 & 38.23 
         & 70.44 & 35.03 & 0.87 
         & 91.42 & 69.51 & 33.56 \\
\rowcolor{gray!10} \rule{0in}{1em}
+ ImagePG (Ours) & 94.97 & 85.99 & 42.56 
                & 80.79 & 43.65 & 3.20 
                & 92.76 & 74.89 & 40.60 \\
\rowcolor{gray!10}
{\tiny \textit{}} 
& {\tiny (\textcolor{blue}{0.15\%↓})} & {\tiny (\textcolor{red}{1.59\%↑})} & {\tiny (\textcolor{red}{4.33\%↑})}
& {\tiny (\textcolor{red}{10.35\%↑})} & {\tiny (\textcolor{red}{8.62\%↑})} & {\tiny (\textcolor{red}{2.33\%↑})}
& {\tiny (\textcolor{red}{1.34\%↑})} & {\tiny (\textcolor{red}{5.38\%↑})} & {\tiny (\textcolor{red}{7.04\%↑})} \\
\bottomrule
\end{tabular}
}
\vspace{-0.75em}
\caption{3D detection performance comparison in terms of different distance ranges (in meters), i.e., [0,20), [20,40), and [40,inf), on the KITTI~\cite{kitti} \textit{val} set calculated by 40 recall positions. $^\dagger$: reproduced.}
\label{tab:kitti_val_dist}
\vspace{-0.5em}
\end{table*}

\renewcommand{\arraystretch}{0.7}
\setlength{\tabcolsep}{12pt}
\begin{table*}[t!]
\centering
\scriptsize
{
\begin{tabular}{l|c c c|c c c|c c c}
\toprule
    \multirow{2}{*}{\textbf{Method}} & \multicolumn{3}{c|}{\textbf{Vehicle 3D} \scriptsize{(IoU=0.7)}} & \multicolumn{3}{c|}{\textbf{Pedestrian 3D} \scriptsize{(IoU=0.5)}} & \multicolumn{3}{c}{\textbf{Cyclist 3D} \scriptsize{(IoU=0.5)}} \\
    \cline{2-10}
    \rule{0pt}{1.25em} & \textbf{[0, 30)} & \textbf{[30, 50)} & \textbf{[50, inf)} & \textbf{[0, 30)} & \textbf{[30, 50)} & \textbf{[50, inf)} & \textbf{[0, 30)} & \textbf{[30, 50)} & \textbf{[50, inf)} \\
\midrule

Baseline$^\dagger$~\cite{pgrcnn} & 89.75 & 66.41 & 38.98
         & 70.38 & 60.28 & 41.38
         & 79.16 & 59.54 & 45.35 \\
\rowcolor{gray!10} \rule{0in}{1em} + ImagePG (Ours) & 90.61 & 69.14 & 42.65
         & 74.22 & 63.95 & 44.98
         & 78.18 & 62.64 & 47.65 \\
\rowcolor{gray!10} \tiny \textit{} 
& \tiny (\textcolor{red}{0.86\%↑}) 
& \tiny (\textcolor{red}{2.73\%↑}) 
& \tiny (\textcolor{red}{3.67\%↑})
& \tiny (\textcolor{red}{3.84\%↑}) 
& \tiny (\textcolor{red}{3.67\%↑}) 
& \tiny (\textcolor{red}{3.60\%↑})
& \tiny (\textcolor{blue}{0.98\%↓}) 
& \tiny (\textcolor{red}{3.10\%↑}) 
& \tiny (\textcolor{red}{2.30\%↑}) \\
\bottomrule
\end{tabular}
}
\vspace{-0.75em}
\caption{3D detection performance (APH) comparison in terms of different distance ranges (in meters), i.e., [0,30), [30,50), and [50,inf), on the Waymo~\cite{waymo} \textit{val} set for difficulty Level 2. $^\dagger$: reproduced.}
\label{tab:waymo_val_dist}
\vspace{-1.5em}
\end{table*}

\subsection{Quantitative Analysis}

\myparagraph{Evaluation on the KITTI and Waymo Datasets.} 
We evaluate our method on the KITTI~\cite{kitti} validation and test sets, as shown in Table~\ref{tab:kitti_val} and Table~\ref{tab:kitti_test}, using PG-RCNN~\cite{pgrcnn} as the reproduced baseline. Our method shows substantial improvements, especially for small and non-rigid classes: +12.01\%p (pedestrian) and +6.47\%p (cyclist) on the validation set, and +1.38\%p (car), +7.91\%p (pedestrian), +5.21\%p (cyclist) on the test set—achieving state-of-the-art cyclist performance on the KITTI leaderboard. To ensure fairness in evaluation, we compare only methods that report all three classes on the test set. In Table~\ref{tab:waymo_val}, we report results on the Waymo~\cite{waymo} dataset, where our method outperforms the baseline across all classes in both mAP and mAPH. 

As shown in Table~\ref{tab:kitti_val_dist} and Table~\ref{tab:waymo_val_dist}, our method exhibits substantial performance improvements across various distance ranges. This consistent trend highlights the effectiveness of image-guided pseudo point generation in addressing the inherent sparsity of LiDAR data at long ranges. By incorporating semantically meaningful visual cues, our method is able to generate informative pseudo points even in regions with limited LiDAR returns, thereby significantly enhancing detection performance in challenging scenarios such as long-range pedestrian and cyclist detection.

\myparagraph{Compatibility with Other Backbones.}
Our proposed approach is built upon a Voxel R-CNN~\cite{voxelrcnn} backbone, yet it is designed to be modular and readily integrable into other 3D object detection frameworks. As demonstrated in Table~\ref{tab:kitti_val_plugin}, we validate its generality by incorporating it into two representative detectors: PV-RCNN~\cite{pvrcnn}, a LiDAR-only model, and UPIDet~\cite{upidet}, a LiDAR–camera fusion-based detector. In both cases, consistent improvements in detection accuracy are observed, demonstrating the broad compatibility and practical utility of our image-guided pseudo point generation. The consistent performance gains observed across diverse backbones highlight the robustness of our method as a universally applicable augmentation module.

\begin{figure*}[t!]
  \centering
  \includegraphics[width=1.0\linewidth]{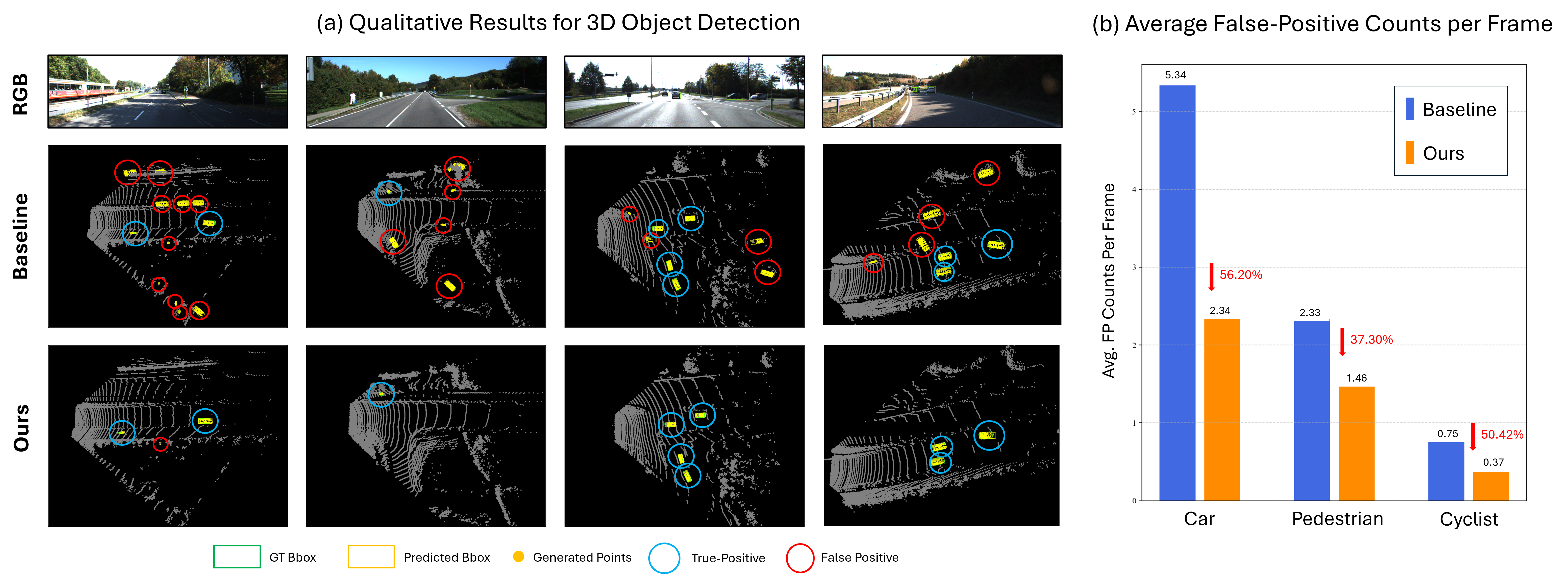}
  \vspace{-2em}
  \caption{Qualitative comparison between the baseline~\cite{pgrcnn} and our method on the KITTI~\cite{kitti} \textit{val} set. Our approach reduces false positives and improves detection performance, whereas the baseline struggles in these scenarios.}
  \label{fig:qualittive_kitti_val}
  \vspace{-1.3em}
\end{figure*}

\myparagraph{Effectiveness in Mitigating False Positives.}  
Figure~\ref{fig:qualittive_kitti_val}(b) illustrates that our method substantially reduces false-positive detections. Specifically, by leveraging semantic cues extracted from the image domain, nearly 50\% of false positives are eliminated compared to the baseline~\cite{pgrcnn}. From a practical perspective, minimizing false positives is crucial for real-world deployment in autonomous driving systems, as it directly reduces the likelihood of misjudging surrounding environments and thus mitigates potential safety hazards. The evaluation is conducted on the KITTI~\cite{kitti} validation set, with predictions filtered by a confidence score threshold of $\geq$ 0.3 to ensure consistency and relevance in quantitative assessment.

\renewcommand{\arraystretch}{1.1}
\begin{table}[t!]
\centering
\scriptsize
\setlength{\tabcolsep}{5.5pt}
{
\begin{tabular}{l||c|c|c c c}
\toprule
    \multirow{2}[1]{*}{\textbf{Method}} & \multirow{2}[1]{*}{\textbf{Modality}} & \multirow{2}[1]{*}{\textbf{mAP}} & \multicolumn{3}{c}{\textbf{Cyclist 3D}} \\
\cline{4-6}
        \rule{0pt}{1.25em} & & & \textbf{Easy} & \textbf{Mod.} & \textbf{Hard} \\
\midrule

        PV-RCNN~\cite{pvrcnn} & L & 73.92 & 86.60 & 69.73 & 65.44 \\
        \rowcolor{gray!10}+ ImagePG (Ours) & L+C & 75.75 & 88.43 & 71.47 & 67.34 \\
        \rowcolor{gray!10} {\tiny \textit{}} & & {\tiny (\textcolor{red}{1.83\%↑})} & {\tiny (\textcolor{red}{1.83\%↑})} & {\tiny (\textcolor{red}{1.74\%↑})} & {\tiny (\textcolor{red}{1.90\%↑})} \\ 

\midrule
        UPIDet~\cite{upidet} & L+C & 80.35 & 94.86 & 74.66 & 71.54 \\
        \rowcolor{gray!10}+ ImagePG (Ours) & L+C & 81.48 & 94.51 & 77.13 & 72.81 \\
        \rowcolor{gray!10} {\tiny \textit{}} & & {\tiny (\textcolor{red}{1.13\%↑})} & {\tiny (\textcolor{blue}{0.35\%↓})} & {\tiny (\textcolor{red}{2.47\%↑})} & {\tiny (\textcolor{red}{1.27\%↑})} \\ 
\bottomrule
\end{tabular}
}
\vspace{-0.5em}
\caption{Performance when integrated into different backbones (PV-RCNN~\cite{pvrcnn} and UPIDet~\cite{upidet}) on the KITTI~\cite{kitti} \textit{val} set.}
\label{tab:kitti_val_plugin}
\vspace{-1.5em}
\end{table}

\subsection{Qualitative Analysis} 
Figure~\ref{fig:qual_point_generation} shows that ImagePG generates points more precisely aligned with ground-truth objects than the baseline~\cite{pgrcnn} and can synthesize points for occluded objects where the baseline often fails. In Figure~\ref{fig:qualittive_kitti_val}(a), qualitative comparisons indicate that the baseline frequently yields false positives—likely due to the lack of dense semantic priors—whereas our method, leveraging visual cues, reduces false positives and achieves robust detection of foreground targets. Qualitative results visualize generated points and predicted boxes with confidence $\geq 0.3$. Additional examples are provided in the supplementary material.

\subsection{Efficiency Analysis}
The inference speed of our model is influenced by the number of refinement stages $N_t$. As shown in Table~\ref{tab:efficiency}, decreasing $N_t$ from 6 to 2 improves throughput to 7.1 frames per second (FPS) while preserving competitive accuracy. It is worth noting that fusion-based detectors are inherently slower than LiDAR-only models, as the incorporation of image backbones introduces extra computational overhead. Nevertheless, our results demonstrate that ImagePG remains sufficiently efficient for real-world deployment. Additional efficiency analysis is provided in the supplementary material.

\subsection{Ablation Studies}

\myparagraph{Effect of each component.} We further evaluate the contribution of each individual module in our framework: I-OPN, MR and IG-RPG modules. As shown in Table~\ref{tab:abl_components_kitti_val}, all modules contribute to enhancing detection performance, with particularly notable improvements for pedestrian and cyclist classes. (i) The IG-RPG module improves detection, achieving results that are competitive with state-of-the-art methods. (ii) The inclusion of the MR module leads to significant performance gains, demonstrating its effectiveness in refining points generation and bounding box predictions. (iii) The full model configuration, incorporating all modules, achieves the highest overall performance across all object categories. These results highlight the effectiveness of our approach in both semantic enrichment and spatial refinement, validating the advantage of integrating all components within a unified framework.


\renewcommand{\arraystretch}{1}
\setlength{\tabcolsep}{5.5pt}
\begin{table}[t!]
\centering
\scriptsize
{
\begin{tabular}{l||c|c|ccc}
\toprule
    \textbf{Method} & \textbf{Modality} & \textbf{FPS} & \textbf{Car 3D} & \textbf{Ped. 3D} & \textbf{Cyc. 3D} \\
\midrule
    PV-RCNN~\cite{pvrcnn} & L & 8.2 & 86.33 & 55.69 & 75.17 \\
    Voxel R-CNN~\cite{voxelrcnn} & L & 16.2 & 86.60 & - & - \\
    PG-RCNN~\cite{pgrcnn} & L & 17.6 & 86.87 & 58.48 & 79.65 \\
\midrule
    LoGoNet~\cite{logonet} & L+C & 9.4 & 87.16 & 64.72 & 79.77 \\
    TED-M~\cite{ted} & L+C & 4.7 & 90.51 & - & - \\
    UPIDet~\cite{upidet} & L+C & 6.0 & 87.32 & 69.64 & 81.35 \\
\midrule
    \rowcolor{gray!10} Ours {\small(\(N_t=2\))} & L+C & 7.1 & 86.90 & 70.62 & 82.15 \\
    \rowcolor{gray!10} Ours {\small(\(N_t=6\))} & L+C & 3.2 & 87.18 & 71.40 & 83.42 \\
\bottomrule
\end{tabular}
}
\vspace{-0.7em}
\captionof{table}{Computational comparison for the KITTI~\cite{kitti} \textit{val} set. All methods are evaluated with single NVIDIA A6000 GPU. 
}
\label{tab:efficiency}
\vspace{-1.7em}
\end{table}

\begin{table}[t!]
\centering
\scriptsize
\setlength{\tabcolsep}{2pt}
{
\begin{tabular}{l||c c c c|c|c c c}
\toprule
    \rule{0pt}{5pt}\textbf{Setting} & \textbf{I-OPN} & \textbf{MR} & \textbf{IG-RPG} & \textbf{RPG} & \textbf{mAP} & \textbf{Car 3D} & \textbf{Ped. 3D} & \textbf{Cyc. 3D} \\
\midrule
        Baseline$^\dagger$ \cite{pgrcnn} & & & & $\checkmark$ & 74.31 & 86.59 & 59.39 & 76.95 \\
\midrule
        (A) & & & $\checkmark$ & & 76.25 & 87.27 & 63.80 & 77.68 \\
        (B) & & $\checkmark$ & & $\checkmark$ & 78.33 & 87.61 & 67.11 & 80.29 \\
        (C) & & $\checkmark$ & $\checkmark$ & & 79.90 & 87.64 & 70.72 & 81.34 \\
        \rowcolor{gray!10} (D) Ours & $\checkmark$ &$\checkmark$  & $\checkmark$ & & 80.67 & 87.18 & 71.40 & 83.42 \\
\bottomrule
\end{tabular}
}
\vspace{-0.7em}
\caption{Effect of each component for 3D detection on the KITTI~\cite{kitti} \textit{val} set. $\dagger$: reproduced.}
\label{tab:abl_components_kitti_val}
\vspace{-1em}
\end{table}

\begin{table}[h]
\label{tab:abl_components}
\centering
\scriptsize
\setlength{\tabcolsep}{6.25pt}
{
\begin{tabular}{l||c c|c|c c c}
\toprule
    \rule{0pt}{5pt}\textbf{Setting} & \textbf{OPN} & \textbf{I-OPN} & \textbf{mAP} & \textbf{Car 3D} & \textbf{Ped. 3D} & \textbf{Cyc. 3D} \\
\midrule
        (A) & & & 79.90 & 87.64 & 70.72 & 81.34 \\
        (B) & $\checkmark$ & & 78.44 & 87.91 & 68.19 & 79.21 \\
        \rowcolor{gray!10} (C) Ours & & $\checkmark$ & 80.67 & 87.18 & 71.40 & 83.42 \\
\bottomrule
\end{tabular}
}
\vspace{-0.7em}
\caption{Effect of the I-OPN module on 3D detection performance on the KITTI~\cite{kitti} \textit{val} set. Both MR and IG-RPG modules are employed in these settings.}
\label{tab:abl_opn}
\vspace{-2em}
\end{table}

\myparagraph{Effectiveness of I-OPN.} Table~\ref{tab:abl_opn} presents the effectiveness of the proposed I-OPN module. Unlike the LiDAR-only OPN~\cite{bshdet3d}, which degrades performance for pedestrians and cyclists, I-OPN consistently improves detection accuracy for these classes. This finding suggests that conventional OPN is inadequate for semantic point generation, whereas our I-OPN enhances the robustness of point generation and consequently improves 3D detection performance.

\vspace{-0.5em}
\section{Conclusion}
\vspace{-0.5em}

In this paper, we propose Image-Guided Semantic Pseudo-LiDAR Point Generation (ImagePG), a novel framework that enhances 3D object detection by leveraging RGB image features to generate dense and semantically meaningful 3D points. Our method generalizes well across diverse detector backbones. In extensive experiments on the KITTI and Waymo datasets, ImagePG achieves superior performance, particularly for small and distant objects such as pedestrians and cyclists, and also significantly reduces false positives. Moreover, ImagePG achieves state-of-the-art performance in cyclist detection on the KITTI benchmark. These results highlight the effectiveness of ImagePG as a strategy to enhance the 3D detection capabilities of conventional fusion or point generation methods. Furthermore, we hope that our method will inspire future research, particularly in advancing robust multi‑modal 3D perception.

\section*{Acknowledgements}
This research was funded by CTO division of LG Innotek Co Ltd. Also, This work was supported by Institute of Information \& communications Technology Planning \& Evaluation (IITP) grant funded by the Korea government(MSIT) (RS-2022-II220043, Adaptive Personality for Intelligent Agents, 20\%). This work was supported by Institute of Information \& communications Technology Planning \& Evaluation (IITP) under the artificial intelligence star fellowship support program to nurture the best talents (IITP-2025-RS-2025-02304828, 20\%) grant, ICT Creative Consilience Program (IITP-2025-RS-2020-II201819, 10\%), and ITRC(Information Technology Research Center) grant (IITP-2025-RS-2022-00156295, 20\%) funded by the Korea government(MSIT). 

{
    \small
    \bibliographystyle{ieeenat_fullname}
    \bibliography{main}
}

\maketitlesupplementary
\newpage
\appendix

\renewcommand{\thefigure}{A\arabic{figure}}
\setcounter{figure}{0}
\renewcommand{\thetable}{A\arabic{table}}
\setcounter{table}{0}

\section{Additional Experimental Details}
\label{apdx:experimental_details}

\subsection{Dense Dataset Construction}
\label{apdx:dense_dataset_construction}
\begin{figure}[t!]
\centering
\includegraphics[width=1.0\linewidth]{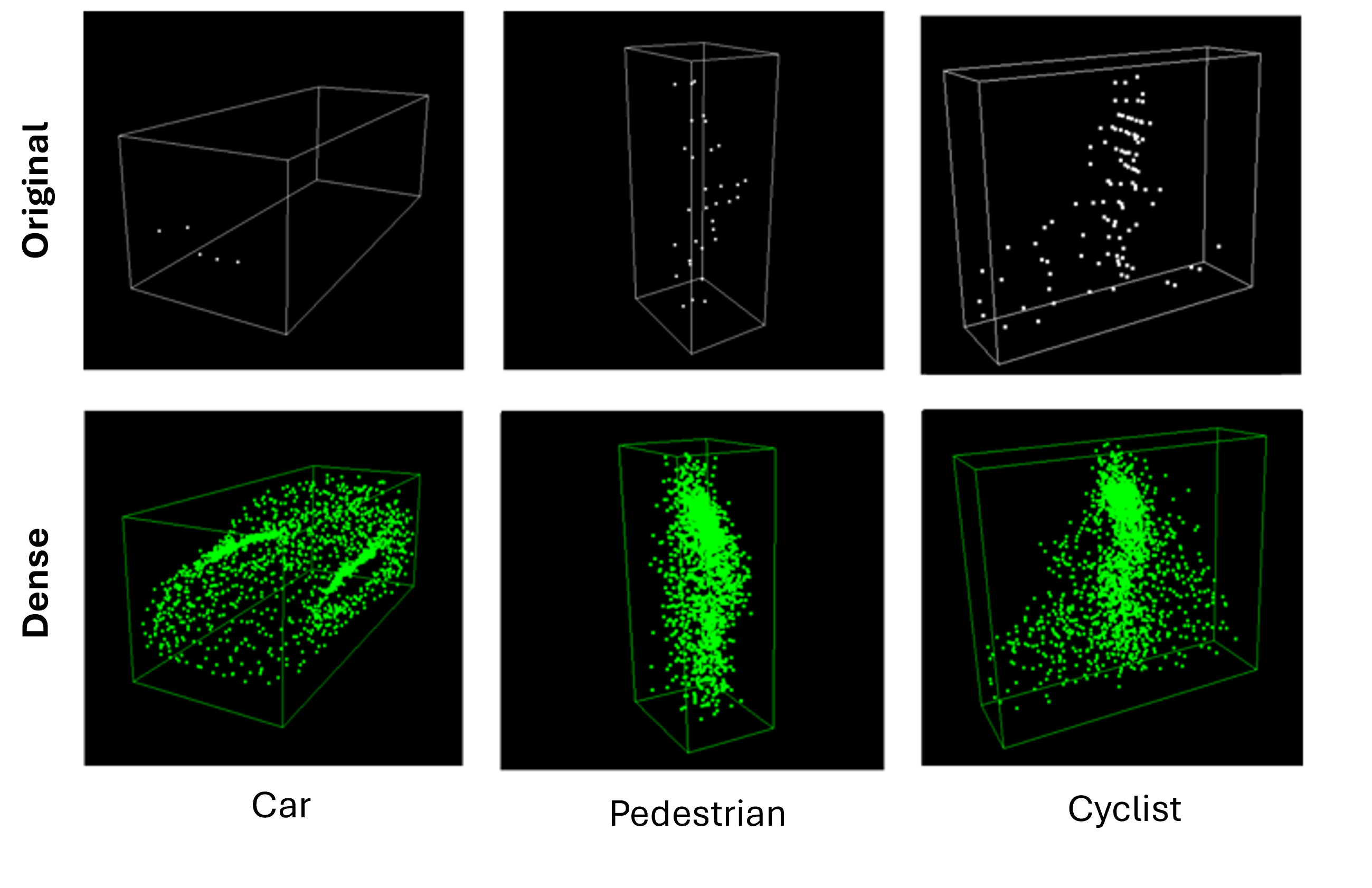}
\vspace{-1.5em}
\caption{Visualization of the dense point cloud dataset for the KITTI~\cite{kitti} benchmark.}
\label{fig:dense_dataset_kitti}
\end{figure}

\begin{figure}[t!]
\centering
\includegraphics[width=1.0\linewidth]{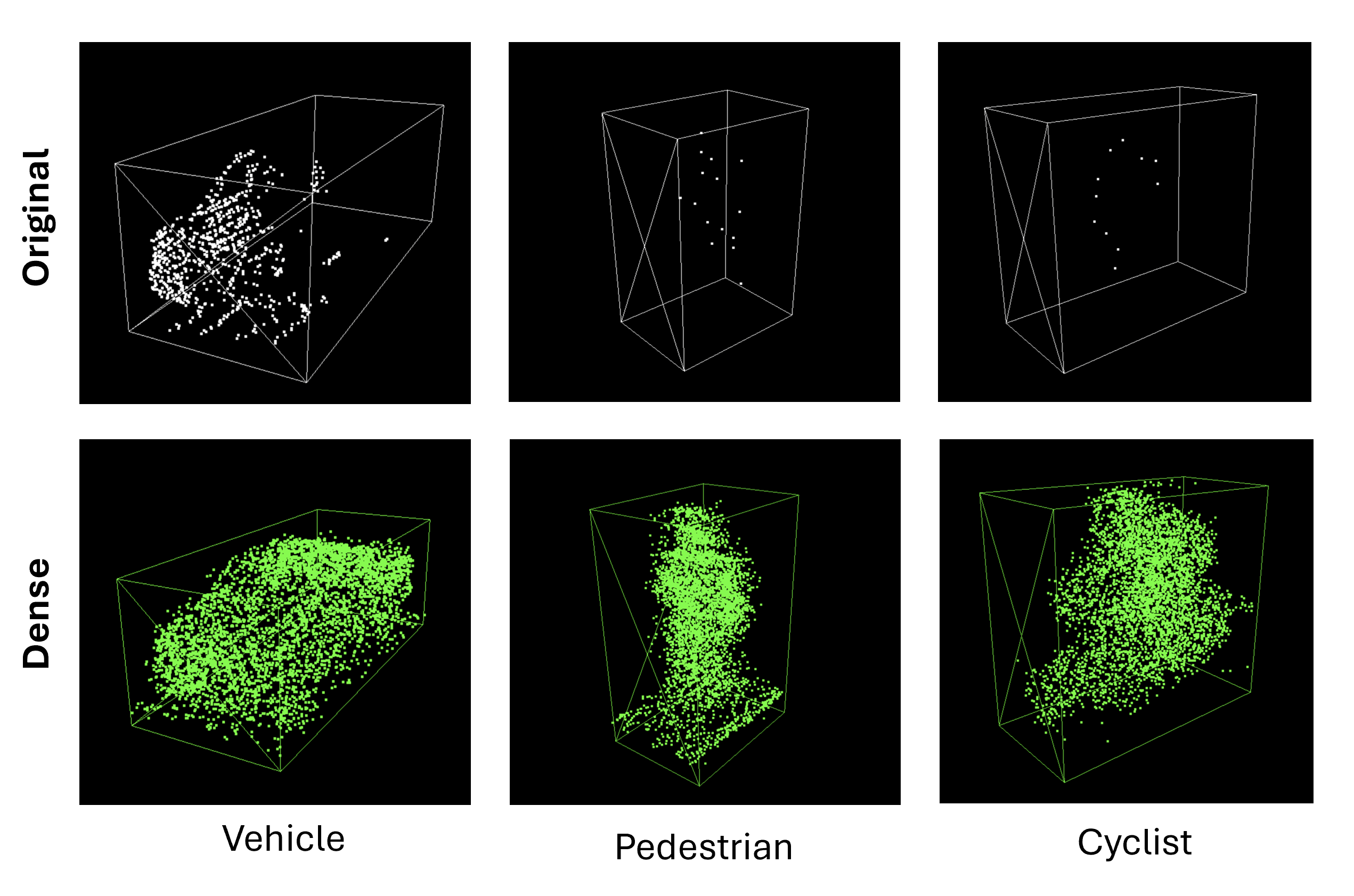}
\vspace{-1.5em}
\caption{Visualization of the dense point cloud dataset for the WOD~\cite{waymo} benchmark.}
\label{fig:dense_dataset_wod}
\end{figure}

To train our network, a dense supervision dataset is required in which ground-truth (GT) objects are represented in a densified or completed form. Following the methodology proposed in BtcDet~\cite{btcdet}, we construct such a dataset using a heuristic, rule-based strategy that combines objects of similar size to compensate for occluded regions. Furthermore, for the car and cyclist classes, we mirror point clouds under the assumption of bilateral symmetry to enhance data completeness. For the Waymo Open Dataset (WOD)~\cite{waymo}, which provides temporal annotations enabling object tracking across frames, we aggregate points from multiple frames corresponding to the same object instance to generate dense representations. Similarly, we assume symmetry for the vehicle and cyclist classes and apply mirroring to further enrich the training data. However, such aggregation methods based on temporal frames are not sufficient to construct complete object shapes, as some parts remain invisible even when multiple frames are aggregated. Therefore, we adopt the approach proposed by BtcDet~\cite{btcdet}, which is used in the KITTI~\cite{kitti}, to construct fully densified object shapes for the Waymo Open Dataset (WOD)~\cite{waymo}. Figure~\ref{fig:dense_dataset_kitti} and ~\ref{fig:dense_dataset_wod} show the dense dataset construction results for the KITTI~\cite{kitti} and WOD~\cite{waymo} benchmarks.

\subsection{Implementation Details}
\label{apdx:implementation_details}

We apply multiple geometric transformations to the input point clouds to enhance transformation-equivariant feature learning. The number of transformation actions \( N_t \) is set to 4 during training and 6 during evaluation for the KITTI~\cite{kitti} dataset. Due to computational cost and VRAM usage for training the whole network, \( N_t \) is set to 3 for both training and evaluation on the Waymo Open Dataset (WOD)~\cite{waymo}. Each transformation action is composed of \texttt{Flip-X}, \texttt{Rotation}, and \texttt{Scaling}. \texttt{Flip-Y} is additionally used for the WOD~\cite{waymo}. Note that the final transformation action preserves the original point cloud distribution, i.e., no flipping, rotation, or scaling is applied for the last transformation action. 

The 2D image backbone employed in ImagePG is Swin Transformer Tiny~\cite{swin}, implemented via OpenPCDet~\cite{openpcdet, bevfusion}. It is pretrained with Mask R-CNN~\cite{maskrcnn} on the nuImages~\cite{nuscenes} dataset for 2D object detection and segmentation, enabling the model to leverage prior knowledge of driving scenes~\cite{bevfusion}. To capture fine-grained features across multiple spatial resolutions, we additionally incorporate a Feature Pyramid Network (FPN)~\cite{fpn}, which aggregates multi-scale features from low-resolution to high-resolution representations. Specifically, we utilize the final high-resolution feature map from the FPN as the image feature input for the subsequent modules. To preserve the pretrained semantic knowledge, we freeze the bottom three layers of the four-layer image backbone and enable gradient backpropagation only through the top layer. 

We employ Anchor Head~\cite{openpcdet} as the region proposal network (RPN) for the KITTI~\cite{kitti} dataset and CenterPoint~\cite{centerpoint} for the Waymo Open Dataset (WOD)~\cite{waymo}. For the IG-RPG module, we utilize a single Transformer Encoder~\cite{transformer} with a hidden dimension of 512. \(N_s\) is set to 4096. RoI grid size ~\({G}\) is set to 6 for convention~\cite{pvrcnn, voxelrcnn}. Adam optimizer~\cite{adam} with a one-cycle policy is used. For the KITTI~\cite{kitti}, we trained for 80 epochs with an initial learning rate of 0.01. We used 8 NVIDIA RTX A6000 GPUs for a batch size of 16, and the training time was less than 12 hours. For the WOD~\cite{waymo}, we trained for 30 epochs with an initial learning rate of 0.01.

Deformable Attention is widely used for 2D and 3D object detection~\cite{deformabledetr, logonet, bevformer} with the model to focus on semantically meaningful regions, even under geometric transformations or sparsity. The deformable attention process can be formulated as follows:
\begin{equation}
\begin{aligned}
    \text{DeformAttn}(Q_i, p_i, x) = \\
    \sum_{m=1}^M W_m \left[ \sum_{k=1}^K A_{mik} \cdot W'_m \bigl \langle x\cdot(p_i + \Delta p_{mik})\bigr \rangle \right],  
\end{aligned}
\end{equation}
where \(p_i\), \(\Delta p_{mik}\), \(A_{mik}\), and \(x\) denote the projected point coordinates, sampling offsets, attention weights of the \(k\)th sampling point in the \(m\)-th attention head and image feature map extracted from the 2D image backbone, respectively. \(W_m\) and \(W'_m\) are learnable weights, and \(\bigl \langle a\cdot b\bigr \rangle\) is a bilinear interpolation of feature map \(a\) from the reference point \(b\). We set the number of attention heads to 4 and sample 4 offset locations for both the I-OPN and IG-RPG modules.


\begin{table}[t!]
\centering
\vspace{0em}
\footnotesize
\setlength{\tabcolsep}{10pt}
{
\begin{tabular}{l||ccccc}
\toprule
    \textbf{Class} & \textbf{Min} & \textbf{Q1} & \textbf{Q2} & \textbf{Q3} & \textbf{Max} \\
\midrule
    Car & 0 & 34 & 111 & 377 & 3573 \\
    Pedestrian & 0 & 54 & 118 & 289 & 1171 \\
    Cyclist & 0 & 25 & 47 & 131 & 1137 \\
\bottomrule
\end{tabular}
}
\captionof{table}{Number of points summarization of the KITTI~\cite{kitti} \textit{val} set.}
\label{tab:kitti_pts_summarize}
\end{table}

\section{Additional Experiments Analysis}
\label{apdx:additional_experiments}

\myparagraph{Effectiveness in Detecting Sparse Object.} We summarize the number of LiDAR points contained within ground truth (GT) bounding boxes in Table~\ref{tab:kitti_pts_summarize}. Similar to object size statistics, a criterion is required to define whether an object is considered sparse. In this study, we define sparse objects as those whose point counts fall within the \(Q_1\) or \(Q_2\) quantiles. Table~\ref{tab:kitti_val_sparse} presents detection performance stratified by point counts within GT bounding boxes. As the results indicate, performance improvements are observed across all categories, with particularly notable gains for the pedestrian and cyclist classes. Specifically, ImagePG improves performance for objects below \(Q_1\) by +2.65\%p and +1.92\%p, and below \(Q_2\) by +15.68\%p and +5.39\%p, respectively. These object types typically exhibit low point densities, underscoring the benefit of our image-guided point generation framework. The results suggest that ImagePG effectively enhances 3D object detection for sparse objects by generating dense and semantically meaningful points informed by image features.

\setlength{\tabcolsep}{1pt}
\renewcommand{\arraystretch}{1.2}
\begin{table*}[h!]
\centering
\footnotesize
\resizebox{1\linewidth}{!}
{
\begin{tabular}{l||c c c c|c c c c|c c c c}
\toprule
    \multirow{2}{*}{\textbf{Method}} & \multicolumn{4}{c|}{\textbf{Car 3D} \scriptsize{(IoU=0.7)}} & \multicolumn{4}{c|}{\textbf{Pedestrian 3D} \scriptsize{(IoU=0.5)}} & \multicolumn{4}{c}{\textbf{Cyclist 3D} \scriptsize{(IoU=0.5)}} \\
    \cline{2-13}
    \rule{0pt}{1em} 
    & \textbf{[Min, Q1)} & \textbf{[Q1, Q2)} & \textbf{[Q2, Q3)} & \textbf{[Q3, Max]} 
    & \textbf{[Min, Q1)} & \textbf{[Q1, Q2)} & \textbf{[Q2, Q3)} & \textbf{[Q3, Max]} 
    & \textbf{[Min, Q1)} & \textbf{[Q1, Q2)} & \textbf{[Q2, Q3)} & \textbf{[Q3, Max]} \\
\midrule
    Baseline~\cite{pgrcnn} 
    & 15.59 & 66.69 & 87.14 & 95.74 
    & 6.34 & 28.05 & 37.82 & 59.96 
    & 7.34 & 35.36 & 77.35 & 86.13 \\ 
\rowcolor{gray!10} \rule{0pt}{1em} 
     + ImagePG (Ours) 
    & 19.34 & 69.38 & 88.56 & 96.07 
    & 8.99 & 43.73 & 57.62 & 78.40 
    & 9.26 & 40.75 & 80.19 & 90.70 \\
\noalign{\vskip -1pt}
\rowcolor{gray!10} \scriptsize \textit{}
    & {\tiny (\textcolor{red}{3.75\%↑})}
    & {\tiny (\textcolor{red}{2.69\%↑})}
    & {\tiny (\textcolor{red}{1.42\%↑})}
    & {\tiny (\textcolor{red}{0.34\%↑})}
    & {\tiny (\textcolor{red}{2.65\%↑})}
    & {\tiny (\textcolor{red}{15.68\%↑})}
    & {\tiny (\textcolor{red}{19.80\%↑})}
    & {\tiny (\textcolor{red}{18.44\%↑})}
    & {\tiny (\textcolor{red}{1.92\%↑})}
    & {\tiny (\textcolor{red}{5.39\%↑})}
    & {\tiny (\textcolor{red}{2.84\%↑})}
    & {\tiny (\textcolor{red}{4.57\%↑})} \\
\bottomrule
\end{tabular}
}
\caption{
Detection performance comparison across number of points quantiles on the KITTI~\cite{kitti} \textit{val} set. 
We use 40 recall positions to compute AP. Objects in lower quantiles (e.g., \(Q_1\), \(Q_2\)) are sparse and more challenging to detect.
}
\label{tab:kitti_val_sparse}
\end{table*}

\setlength{\tabcolsep}{11pt}
\renewcommand{\arraystretch}{1}
\begin{table*}[h!]
\centering
\footnotesize
\resizebox{1\textwidth}{!}
{
\begin{tabular}{l||c c c|c c c|c c c}
\toprule
    \multirow{2}{*}{\textbf{Method}} & \multicolumn{3}{c|}{\textbf{Car 3D} \scriptsize{(IoU=0.7)}} & \multicolumn{3}{c|}{\textbf{Pedestrian 3D} \scriptsize{(IoU=0.5)}} & \multicolumn{3}{c}{\textbf{Cyclist 3D} \scriptsize{(IoU=0.5)}} \\
    \cline{2-10}
    \rule{0pt}{1.25em} 
    & \textbf{LVL\_1} & \textbf{LVL\_2} & \textbf{LVL\_3} 
    & \textbf{LVL\_1} & \textbf{LVL\_2} & \textbf{LVL\_3} 
    & \textbf{LVL\_1} & \textbf{LVL\_2} & \textbf{LVL\_3} \\
\midrule
    Baseline~\cite{pgrcnn} 
    & 84.62 & 77.01 & 54.41 
    & 60.48 & 19.56 & 4.49 
    & 83.50 & 25.60 & 0.93 \\
\rowcolor{gray!10} \rule{0pt}{1em} 
     + ImagePG (Ours) 
    & 84.77 & 79.58 & 60.47 
    & 73.96 & 33.93 & 9.02 
    & 85.53 & 28.30 & 2.79 \\
\noalign{\vskip -1pt}
\rowcolor{gray!10} \scriptsize \textit{}
    & {\tiny (\textcolor{red}{0.15\%↑})}
    & {\tiny (\textcolor{red}{2.57\%↑})}
    & {\tiny (\textcolor{red}{6.06\%↑})}
    & {\tiny (\textcolor{red}{13.48\%↑})}
    & {\tiny (\textcolor{red}{14.37\%↑})}
    & {\tiny (\textcolor{red}{4.53\%↑})}
    & {\tiny (\textcolor{red}{2.03\%↑})}
    & {\tiny (\textcolor{red}{2.70\%↑})}
    & {\tiny (\textcolor{red}{1.86\%↑})} \\
\bottomrule
\end{tabular}
}
\caption{
Detection performance comparison across different occlusion levels on the KITTI~\cite{kitti} \textit{val} set.
We use 40 recall positions to compute AP. Higher occlusion levels correspond to more severely occluded objects.
}
\label{tab:kitti_val_occlusion}
\end{table*}

\setlength{\tabcolsep}{8pt}
\renewcommand{\arraystretch}{1.25}
\begin{table*}[t!]
\centering
\footnotesize
{
\begin{tabular}{c||l|c c c|c c c|c c c}
\toprule
    \multirow{2}{*}{\textbf{LiDAR Beams}} & \multirow{2}{*}{\textbf{Method}} 
    & \multicolumn{3}{c|}{\textbf{Car 3D} \scriptsize{(IoU=0.7)}} 
    & \multicolumn{3}{c|}{\textbf{Pedestrian 3D} \scriptsize{(IoU=0.5)}} 
    & \multicolumn{3}{c}{\textbf{Cyclist 3D} \scriptsize{(IoU=0.5)}} \\
    \cline{3-11}
    \rule{0pt}{0em} & & \textbf{Easy} & \textbf{Mod.} & \textbf{Hard} 
                   & \textbf{Easy} & \textbf{Mod.} & \textbf{Hard} 
                   & \textbf{Easy} & \textbf{Mod.} & \textbf{Hard} \\
\midrule

\multirow{3}{*}{\textbf{64}} 
    & Baseline$^\dagger$~\cite{pgrcnn} 
    & 92.49 & 84.87 & 82.40 
    & 65.86 & 58.85 & 53.47 
    & 91.29 & 72.13 & 67.44 \\   
    
    & \cellcolor{gray!10} + ImagePG (Ours) 
    & \cellcolor{gray!10} 92.41 & \cellcolor{gray!10} 85.67 & \cellcolor{gray!10} 83.45 
    & \cellcolor{gray!10} 77.64 & \cellcolor{gray!10} 71.01 & \cellcolor{gray!10} 65.56 
    & \cellcolor{gray!10} 96.53 & \cellcolor{gray!10} 78.94 & \cellcolor{gray!10} 74.79 \\
    
    \noalign{\vskip -1pt}
    & \cellcolor{gray!10} \scriptsize \textit{} 
    & \cellcolor{gray!10} {\tiny (\textcolor{blue}{0.08\%↓})} 
    & \cellcolor{gray!10} {\tiny (\textcolor{red}{0.80\%↑})} 
    & \cellcolor{gray!10} {\tiny (\textcolor{red}{1.05\%↑})}
    & \cellcolor{gray!10} {\tiny (\textcolor{red}{11.78\%↑})} 
    & \cellcolor{gray!10} {\tiny (\textcolor{red}{12.16\%↑})} 
    & \cellcolor{gray!10} {\tiny (\textcolor{red}{12.09\%↑})}
    & \cellcolor{gray!10} {\tiny (\textcolor{red}{5.24\%↑})} 
    & \cellcolor{gray!10} {\tiny (\textcolor{red}{6.81\%↑})} 
    & \cellcolor{gray!10} {\tiny (\textcolor{red}{7.35\%↑})} \\

\midrule

\multirow{3}{*}{\textbf{32}} 
    & Baseline$^\dagger$~\cite{pgrcnn} 
    & 70.59 & 48.97 & 44.26 
    & 62.19 & 54.16 & 48.15 
    & 66.72 & 40.79 & 38.55 \\
    

& \cellcolor{gray!10} + ImagePG (Ours)
& \cellcolor{gray!10} 78.82 & \cellcolor{gray!10} 58.40 & \cellcolor{gray!10} 53.27 
& \cellcolor{gray!10} 72.80 & \cellcolor{gray!10} 65.23 & \cellcolor{gray!10} 58.26 
& \cellcolor{gray!10} 70.81 & \cellcolor{gray!10} 44.91 & \cellcolor{gray!10} 42.51 \\
    
    \noalign{\vskip -1pt}
    & \cellcolor{gray!10} \scriptsize \textit{} 
    & \cellcolor{gray!10} {\tiny (\textcolor{red}{8.23\%↑})} 
    & \cellcolor{gray!10} {\tiny (\textcolor{red}{9.43\%↑})} 
    & \cellcolor{gray!10} {\tiny (\textcolor{red}{9.01\%↑})}
    & \cellcolor{gray!10} {\tiny (\textcolor{red}{10.61\%↑})} 
    & \cellcolor{gray!10} {\tiny (\textcolor{red}{11.07\%↑})} 
    & \cellcolor{gray!10} {\tiny (\textcolor{red}{10.11\%↑})}
    & \cellcolor{gray!10} {\tiny (\textcolor{red}{4.09\%↑})} 
    & \cellcolor{gray!10} {\tiny (\textcolor{red}{4.12\%↑})} 
    & \cellcolor{gray!10} {\tiny (\textcolor{red}{3.96\%↑})} \\

\midrule

\multirow{3}{*}{\textbf{16}} 
    & Baseline$^\dagger$~\cite{pgrcnn} 
    & 52.17 & 33.27 & 28.44 
    & 40.91 & 35.16 & 30.88 
    & 35.70 & 20.39 & 18.72 \\
    
    & \cellcolor{gray!10} + ImagePG (Ours) 
    & \cellcolor{gray!10} 68.10 & \cellcolor{gray!10} 45.38 & \cellcolor{gray!10} 39.15 
    & \cellcolor{gray!10} 60.72 & \cellcolor{gray!10} 52.52 & \cellcolor{gray!10} 46.26 
    & \cellcolor{gray!10} 54.20 & \cellcolor{gray!10} 31.55 & \cellcolor{gray!10} 29.66 \\
    
    \noalign{\vskip -1pt}
    & \cellcolor{gray!10} \scriptsize \textit{} 
    & \cellcolor{gray!10} {\tiny (\textcolor{red}{15.93\%↑})} 
    & \cellcolor{gray!10} {\tiny (\textcolor{red}{12.11\%↑})} 
    & \cellcolor{gray!10} {\tiny (\textcolor{red}{10.71\%↑})}
    & \cellcolor{gray!10} {\tiny (\textcolor{red}{19.81\%↑})} 
    & \cellcolor{gray!10} {\tiny (\textcolor{red}{17.36\%↑})} 
    & \cellcolor{gray!10} {\tiny (\textcolor{red}{15.38\%↑})}
    & \cellcolor{gray!10} {\tiny (\textcolor{red}{18.50\%↑})} 
    & \cellcolor{gray!10} {\tiny (\textcolor{red}{11.16\%↑})} 
    & \cellcolor{gray!10} {\tiny (\textcolor{red}{10.94\%↑})} \\

\bottomrule
\end{tabular}
}
\caption{
Detection performance comparison across different LiDAR specification settings on the KITTI~\cite{kitti} \textit{val} set. 
We use 40 recall positions to compute AP. $\dagger$: Reproduced.
}
\label{tab:kitti_val_lidar_beams}
\vspace{-1em}
\end{table*}

\renewcommand{\arraystretch}{1}
\setlength{\tabcolsep}{5pt}
\begin{table}[t]
\centering
\resizebox{1.0\linewidth}{!}{
\begin{tabular}{c||c|cc|ccc}
\toprule
    \textbf{Setting} & \(\mathbf{N_t}\) & \textbf{FPS} & {\textbf{VRAM}} & \textbf{Car 3D} & \textbf{Ped. 3D} & \textbf{Cyc. 3D} \\
\midrule
    (a) & 2 & 7.1 & 4.0 & 86.90 & 70.62 & 82.15 \\
    (b) & 3 & 5.5 & 4.4 & 87.11 & 71.02 & 81.66 \\
    (c) & 4 & 4.6 & 4.4 & \textbf{87.20} & 70.83 & 82.36 \\
    (d) & 5 & 3.8 & 4.5 & 87.12 & 70.76 & \textbf{84.09} \\
    (e) & 6 & 3.2 & 4.7 & 87.18 & \textbf{71.40} & 83.42 \\
\bottomrule
\end{tabular}
}
\caption{Computational comparison for the KITTI~\cite{kitti} \textit{val} set. Note that evaluation is conducted with a single NVIDIA A6000 GPU.}
\label{tab:refine_mAP}
\end{table}

\renewcommand{\arraystretch}{1}
\setlength{\tabcolsep}{5pt}

\begin{table}[t]
\centering
\resizebox{1.0\linewidth}{!}{
\begin{tabular}{c||ccccc|c}
\toprule
\textbf{Module} & 2D & 3D & I-OPN & RPN & RoI Head & Total \\
\midrule
\textbf{\# Params (M)} & 29.13 & 1.00 & 2.88 & 7.31 & 12.22 & 52.54 \\
\bottomrule
\end{tabular}
}
\vspace{-0.5em}
\caption{Number of parameters of each module in ImagePG.}
\label{tab:params}
\vspace{-1.5em}
\end{table}

\begin{figure}[t!]
  \centering
  \includegraphics[width=0.9\linewidth]{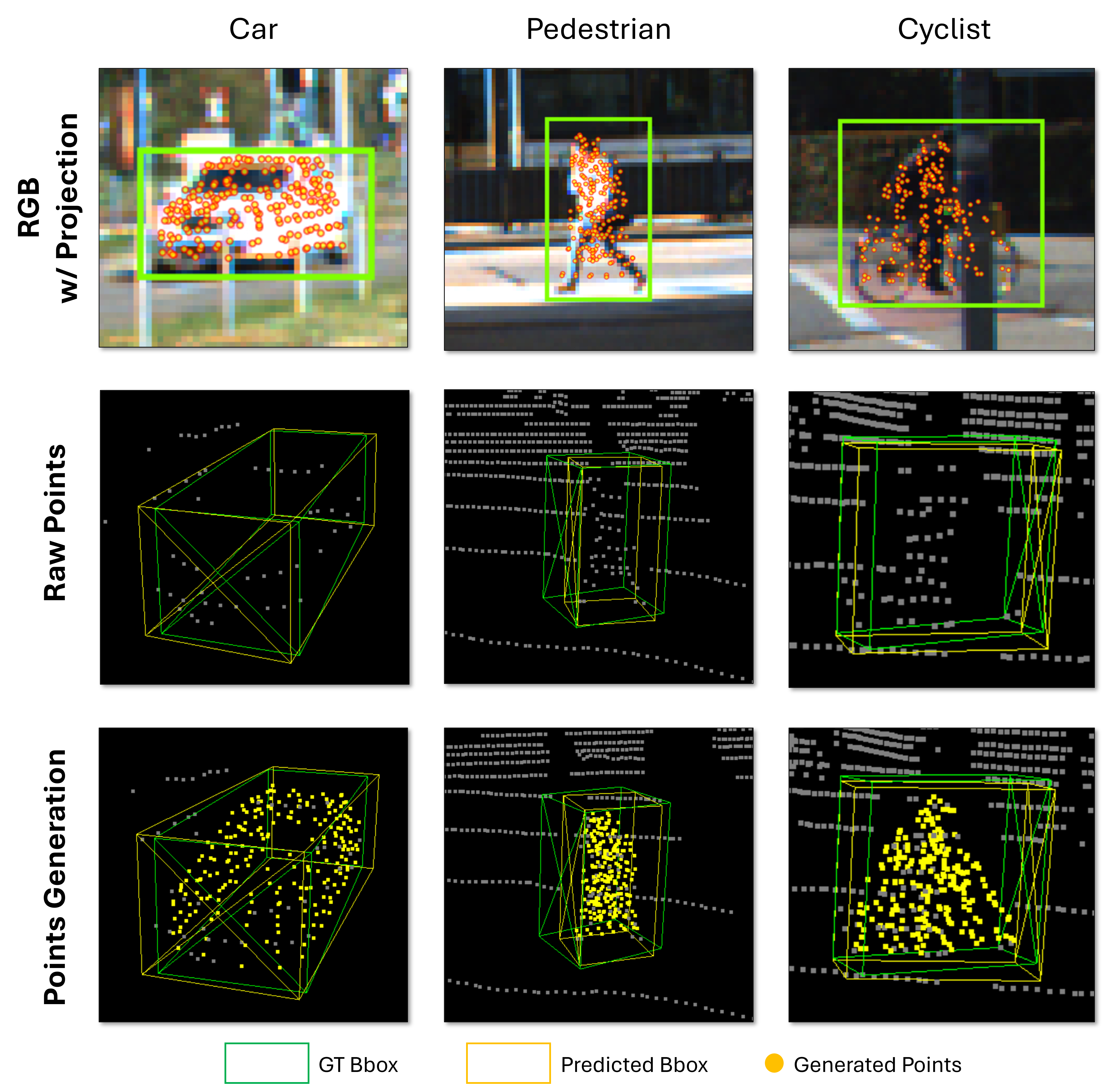}
  \vspace{-0em}
  \caption{Qualitative results of semantic point generation. The generated points align well with image objects.}
  \label{fig:qualitative_kitti_val_pg_compare}
  \vspace{-1em}
\end{figure}

\myparagraph{Effectiveness in Detecting Occluded Object.} Occlusion from surrounding objects or background elements often leads to sparsely sampled point clouds, as occluded regions cannot be captured by LiDAR sensors. To assess the robustness of our proposed ImagePG framework in handling such occlusions, we analyze its performance in object detection under varying occlusion levels. The KITTI~\cite{kitti} benchmark provides occlusion annotations categorized into three levels: 1 (fully visible), 2 (partially occluded), and 3 (heavily occluded). As presented in Table~\ref{tab:kitti_val_occlusion}, ImagePG consistently outperforms the LiDAR-only baseline~\cite{pgrcnn} across all occlusion levels. Notably, our method achieves substantial improvements even for heavily occluded objects (level 3), demonstrating that image-guided point generation enriches semantic context and effectively compensates for LiDAR sparsity caused by occlusion.

\myparagraph{Generalization to LiDAR Resolution.}
LiDAR sensors yield varying point cloud densities depending on hardware specifications. To evaluate the generalizability of ImagePG under different LiDAR configurations—particularly in low-resolution scenarios—we analyze its performance under varying input sparsity levels. Specifically, we simulate reduced beam counts by converting LiDAR scans into range maps and downsampling their vertical resolution to emulate 32-beam and 16-beam sensors. The resulting range maps are projected back into 3D space to generate sparse point clouds. Both ImagePG and the baseline~\cite{pgrcnn} are trained from scratch on these sparsified inputs using the same configuration as the full 64-beam setting of the KITTI~\cite{kitti}.

\begin{figure*}[t]
  \centering
  \includegraphics[width=1.0\linewidth]{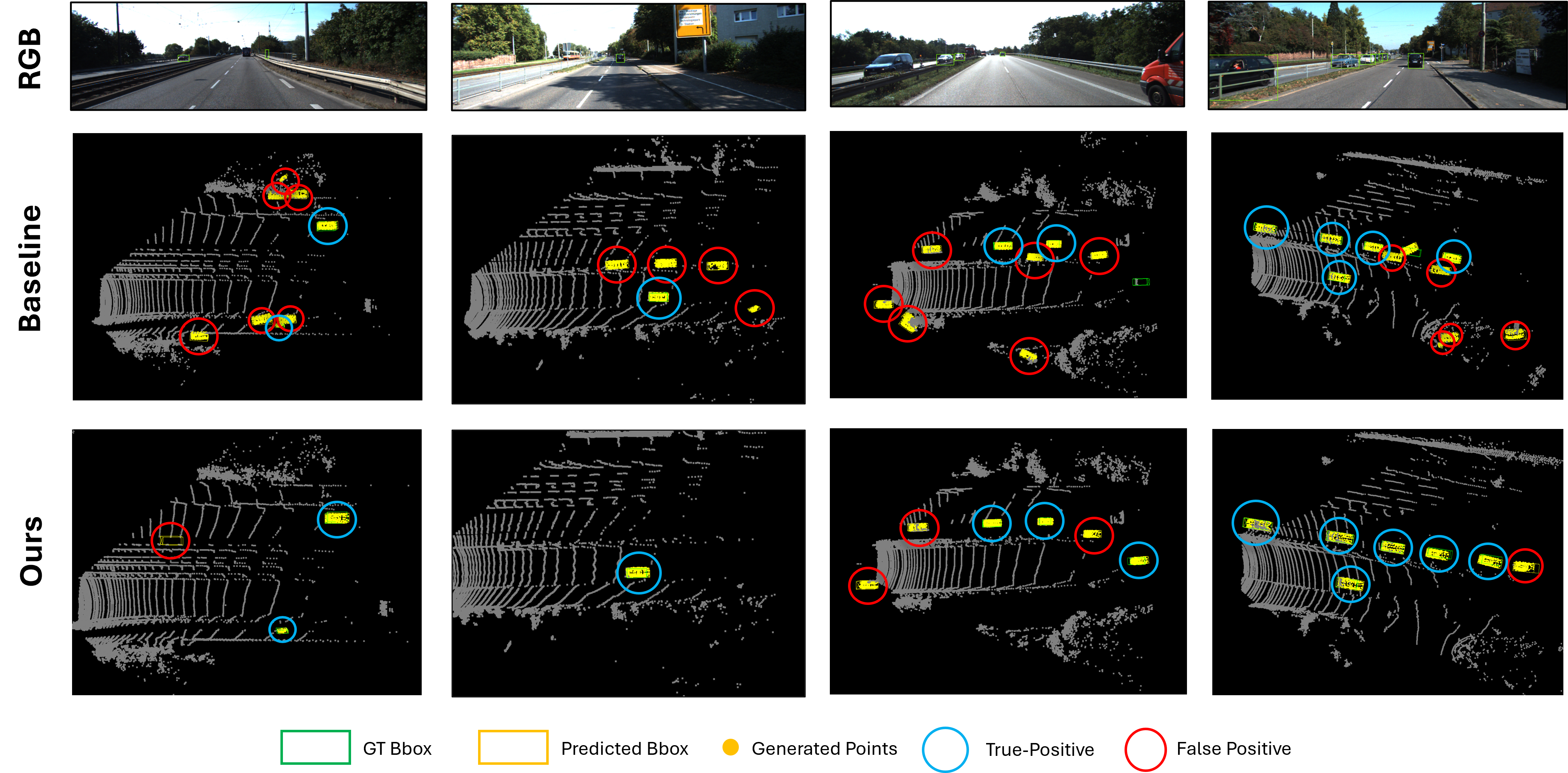}
  \vspace{-1.5em}
  \caption{Additional qualitative comparison between the baseline~\cite{pgrcnn} and our proposed ImagePG on the KITTI~\cite{kitti} \textit{val} set. ImagePG demonstrates well-aligned and semantically meaningful point generation while maintaining a low false positive rate.}
  \label{fig:qualittive_kitti_val_2}
\end{figure*}
\begin{figure*}[h]
  \centering
  \includegraphics[width=1.0\linewidth]{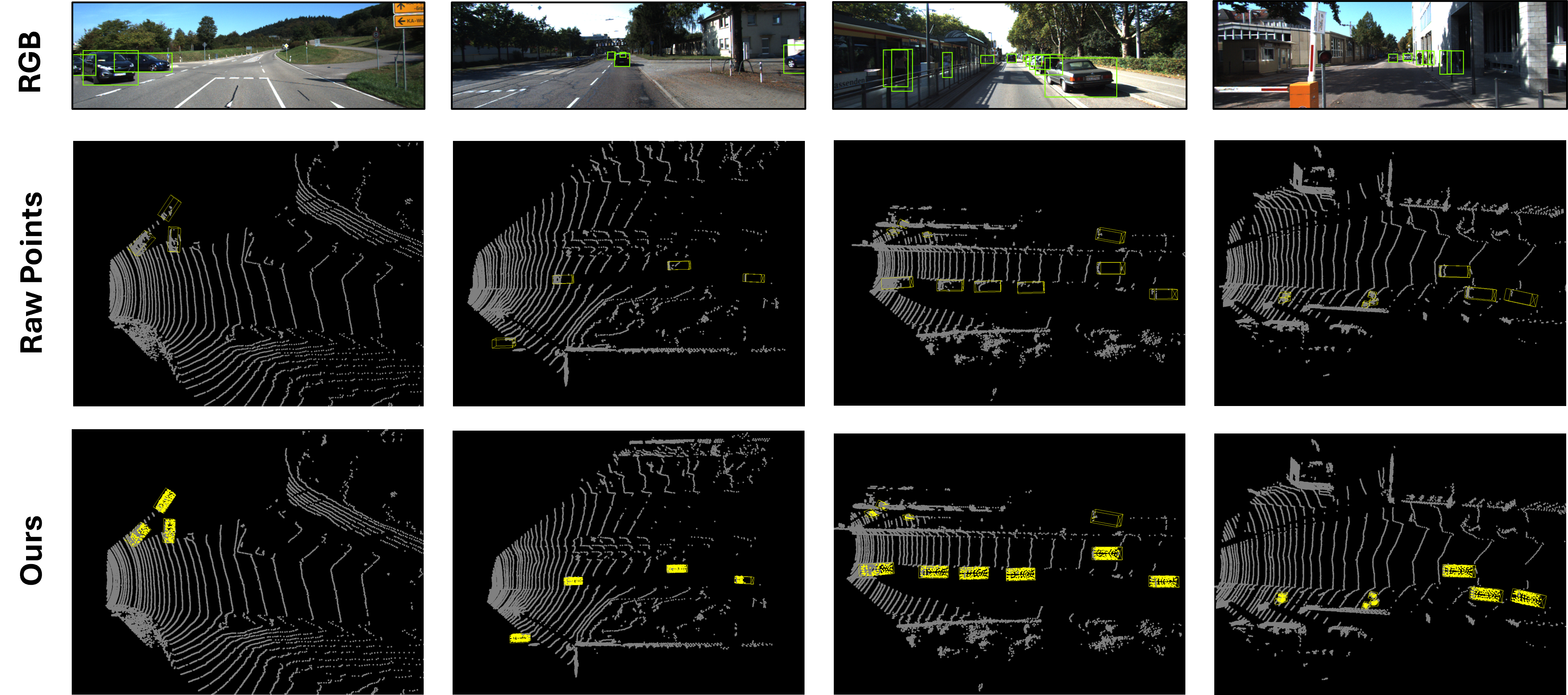}
  \vspace{-1.5em}
  \caption{Detection and point generation results on the KITTI \cite{kitti} \textit{test} set.}
  \label{fig:qualittive_kitti_test}
\end{figure*}

Table~\ref{tab:kitti_val_lidar_beams} presents the AP results across different LiDAR beam densities on the KITTI~\cite{kitti} \textit{val} set. As the number of LiDAR beams decreases, ImagePG exhibits even larger performance improvements over the baseline~\cite{pgrcnn}. These results highlight that ImagePG not only enhances detection accuracy with high-resolution LiDAR but also demonstrates robustness under low-resolution configurations. This makes ImagePG a versatile and practical solution across diverse LiDAR hardware platforms, particularly enabling effective deployment with low-cost, low-resolution LiDAR sensors.

\subsection{Additional Qualitative Analysis}
\label{apdx:additional_qualitative_results}

In Figure~\ref{fig:qualitative_kitti_val_pg_compare}, we visualize the input images, original point clouds, and our generated semantic point clouds. The generated points are well aligned with visual semantics and effectively densify the original LiDAR data. Figure \ref{fig:qualittive_kitti_val_2} and \ref{fig:qualittive_kitti_test} show the additional detection and points generation results for the KITTI~\cite{kitti} validation and test set, respectively. The results indicate that most objects are successfully detected, with high quality points generation, maintaining low false positives. Note that confidence scores larger than 0.3 are only visualized.

\subsection{Additional Efficiency Analysis}
\label{apdx:efficiency_analysis}

Table~\ref{tab:refine_mAP} presents detection performance under varying numbers of transformation actions. For accuracy-critical applications, configuration (e) yields competitive performance, while configuration (a) offers significantly improved inference speed, making it preferable for latency-sensitive deployments. Table~\ref{tab:params} details the parameters, with a total of 52.54M parameters, over half of which (29.13M) are attributed to the image backbone. The overall GPU memory footprint remains moderate at approximately 4.7~GiB, demonstrating the practicality of ImagePG.


\end{document}